\documentclass[lettersize, journal]{IEEEtran} 

\IEEEoverridecommandlockouts                              

\usepackage{graphics} 
\usepackage{epsfig} 
\usepackage{mathptmx} 
\usepackage{times} 
\usepackage{amsmath} 
\usepackage{amssymb}  
\usepackage{xcolor}
\usepackage{subcaption}
\usepackage{graphicx}
\usepackage{booktabs}
\usepackage{bm}
\usepackage{hyperref}
\usepackage{dsfont}
\usepackage{algorithmic}

\DeclareMathOperator*{\argmax}{arg max}

\title{\LARGE \bf
Bayesian Active Object Recognition and 6D Pose Estimation from Multimodal Contact Sensing
}

\author{Haodong Zheng$^{1}$, Gabriele M. Caddeo$^{2}$, Andrei C. Jalba$^{1}$, Wijnand A. IJsselsteijn$^{1}$,\\ Lorenzo Natale$^{2}$, Raymond H. Cuijpers$^{1}$
\thanks{*This work was not supported by any organization}
\thanks{$^{1}$Haodong Zheng, Andrei C. Jalba, Raymond H. Cuijpers and Wijnand A. IJsselsteijn are with Eindhoven University of Technology, 5612 AZ Eindhoven, The Netherlands. 
        {\tt\small e-mails: h.zheng3@tue.nl; a.c.jalba@tue.nl; r.h.cuijpers@tue.nl; w.a.ijsselsteijn@tue.nl}}%
\thanks{$^{2}$Gabriele M. Caddeo and Lorenzo Natale are with the Humanoid Sensing and Perception Group, Italian Institute of Technology,
        16163 Genova GE, Italy
        {\tt\small e-mails: gabriele.caddeo@iit.it; lorenzo.natale@iit.it}}%
}
\begin{document}

\twocolumn[{%
\renewcommand\twocolumn[1][]{#1}%
\maketitle
\begin{center}
    \vspace{-0.1in}
    \centering
    \captionsetup{type=figure}
    \includegraphics[width=\linewidth]{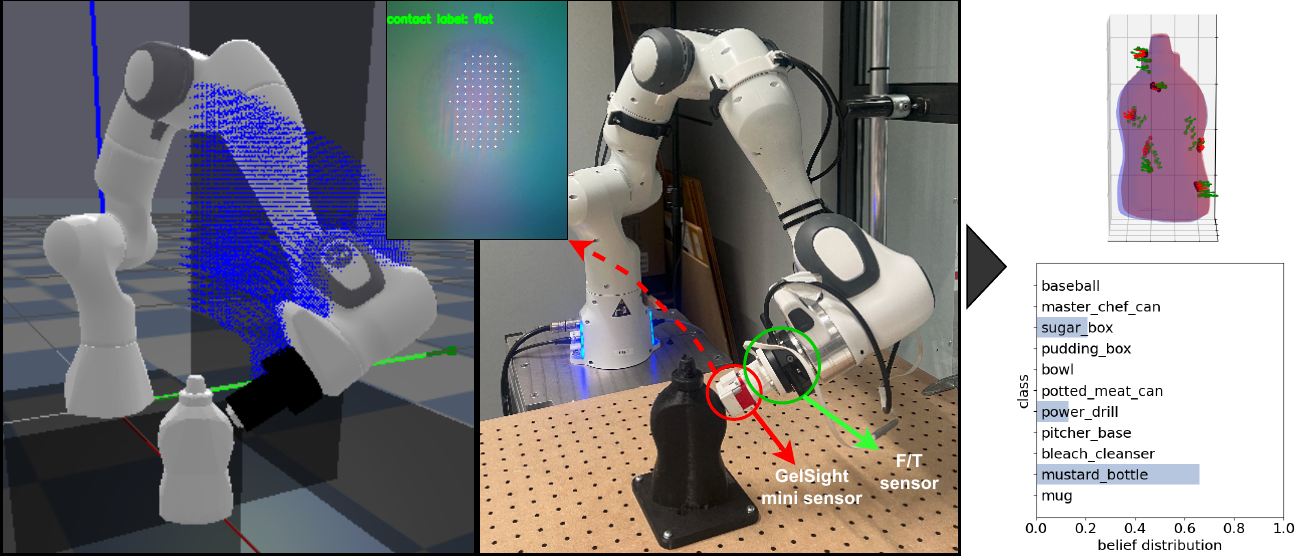}
    \captionof{figure}{Side-by-side comparison showing (left) the calibrated simulated setup, (center) its real-world counterpart, and (right) the corresponding results. Our framework leverages rich information from robot motion, a force torque sensor and GelSight sensor to perform active object recognition and pose estimation within a few touches (actions). The blue points on the simulated setup figure represents the swept volume of the robot arm, the middle figure shows the tactile image from GelSight, where a contact surface type and a contact patch is obtained. The blue shape in the results indicates the ground truth of the object class and pose, while the red shape indicates the MAP estimation of the object class and pose. The points with red normals are the contact points extracted from the GelSight sensor, while the points with green normal vectors are the contact points estimated from the force torque signals.}
    \label{fig:scheme}
    \vspace{-0.1in}
\end{center}
}]

\thispagestyle{empty}
\pagestyle{empty}
\def\BibTeX{{\rm B\kern-.05em{\sc i\kern-.025em b}\kern-.08em
    T\kern-.1667em\lower.7ex\hbox{E}\kern-.125emX}}


\begin{abstract}
 We present an active tactile exploration framework for joint object recognition and 6D pose estimation. The proposed method integrates wrist force/torque sensing, GelSight tactile sensing, and free-space constraints within a Bayesian inference framework that maintains a belief over object class and pose during active tactile exploration. By combining contact and non-contact evidence, the framework reduces ambiguity and improves robustness in the joint class-pose estimation problem. To enable efficient inference in the large hypothesis space, we employ a customized particle filter that progressively samples particles based on new observations. The inferred belief is further used to guide active exploration by selecting informative next touches under reachability constraints. For effective data collection, a motion planning and control framework is developed to plan and execute feasible paths for tactile exploration, handle unexpected contacts and GelSight-surface alignment with tactile servoing. We evaluate the framework in simulation and on a Franka Panda robot using 11 YCB objects. Results show that incorporating tactile and free-space information substantially improves recognition and pose estimation accuracy and stability, while reducing the number of action cycles compared with force/torque-only baselines. Code, dataset, and supplementary material will be made available online.
\end{abstract}

\section{Introduction}

\IEEEPARstart{R}{obots} that operate outside carefully engineered factory settings inevitably have to deal with uncertainty: they often do not know which object they are interacting with, where it is located, or how it is oriented, and visual sensing is frequently compromised by occlusions, specularities, or clutter. In these conditions, physical contact is not merely an exceptional event to be avoided, but a rich source of information. Using touch geometric and material properties can be recognised and object's can be inferred~\cite{klatzkyIdentifyingObjectsTouch1985}. Similar tactile capabilities are highly desirable for robotic platforms, since accurate object identity and pose estimates are critical for grasp planning and dexterous manipulation~\cite{11162616}. 

Yet, despite its potential, contact information is often underused. In many systems, “contact” is treated as a binary signal (contact/no contact)~\cite{Tandem3d_2023} or, at best, as a sparse set of contact points inferred from wrist force/torque (F/T) measurements. Vision-based approaches to object recognition and pose estimation are widely deployed, but their performance can degrade significantly under occlusion and poor lighting conditions~\cite{11128495}. Meanwhile, the emergence of different tactile sensing technologies~\cite{magneticjamone, reviewtactip, gelfinger} provides an alternative source of perception that is largely insensitive to these visual failure modes. In particular, vision-based tactile sensors~\cite{abadVisuotactileSensorsEmphasis2020} such as GelSight~\cite{yuanGelSightHighResolutionRobot2017}, GelSlim~\cite{taylorGelSlimHighResolutionMeasurement2022, gelslim4} and DIGIT~\cite{lambetaDIGITNovelDesign2020, lambeta2024digitizingtouchartificialmultimodal} can recover fine-grained local surface geometry at contact and, in many cases, additional cues such as force-related signals and texture~\cite{shahidzadeh2024feelanyforceestimatingcontactforce,Li_2024}. However, tactile observations are inherently local: each contact patch covers only a small region, and relying on tactile sensing alone makes it challenging to recover global context without actively collecting information across multiple touches.

Crucially, contact does not yield a single type of measurement; instead, it produces a multimodal bundle of signals that encode different aspects of the environment. Firstly, interaction forces and torques describe how the environment resists the robot’s motion and can be used to infer the line of action of contact and assess the plausibility of hypothesized object poses. Second, vision-based tactile sensing provides dense, local geometric information (contact patches, normals, and local surface structure) that is difficult to obtain reliably with external cameras at close range. Third, and often overlooked, the absence of contact is informative: when the robot moves through space without colliding, it acquires “negative” evidence that the object cannot occupy that region. This free-space information can dramatically prune the set of feasible object hypotheses, especially early in exploration when uncertainty is large. Together, force interaction, tactile geometry, and free-space constraints provide complementary evidence that is rarely integrated in a unified perception pipeline, despite its potential to reduce ambiguity efficiently.
Since a single tactile measurement only provides local information, ambiguity has to be resolved by actively taking multiple measurements. As a result, the robot must decide where to touch next, plan safe motions in the presence of uncertain object pose, and maintain controlled contact to obtain stable, informative tactile images.

In this paper, we argue that robust object recognition and 6D pose estimation in the real world requires treating contact as a multimodal sensing event and fusing its complementary signals in a unified inference and control loop. We propose an active tactile exploration framework that combines (i) wrist force/torque measurements, (ii) tactile contact patches from a GelSight-class sensor, including contact-type labels (e.g., planar, edge, corner) that capture local surface structure, and (iii) motion-derived free-space “negative contact” information. At each time step, the tactile sensor provides a contact area and oriented contact observations, and the contact-type label is used to modulate the likelihood (e.g., down-weighting normal consistency near edges/corners where the signed distance field (SDF) gradient is ill-defined). These measurements are integrated into a Bayesian pipeline based on a particle filter that jointly estimates object class and pose. Compared to approaches that rely on a single contact signal, the proposed fusion leverages both positive evidence (where contact occurs) and negative evidence (where the robot has moved without contact) to reduce uncertainty more rapidly.

To make inference tractable in a large class–pose space, we approximate the posterior with a particle-based representation and employ point-pair-feature-based proposal heuristics that progressively concentrate samples around plausible hypotheses as new observations arrive. Based on the current estimate, the framework actively selects the next target point of interest on the maximum a posterior (MAP) hypothesis to acquire new data, while accounting for reachability and previously unreachable targets. We close the loop with a control stack that plans collision-free motions to pre-touch configurations and executes contacts with compliance. In addition, we use tactile servoing to align the tactile sensor with the object surface and center the contact patch on the sensor, improving data quality and robustness under model mismatch. We evaluate the full autonomous pipeline in simulation and on a real Franka Panda robot using 11 YCB~\cite{7254318} objects, and show that multimodal contact fusion enables accurate object recognition and 6D pose estimation with only a small number of action cycles ($4/6$ touches). Fig.~\ref{fig:scheme} shows the simulated and real-world scenarios with an example of MAP estimation of object class and pose. 

The main contributions of this work are threefold. \textbf{First}, we formulate a unified probabilistic framework that combines force-based contact constraints, tactile surface geometry (including contact-type information) and free-space constraints, explicitly leveraging both positive and negative evidence from physical interaction. \textbf{Second}, we present an active exploration and control pipeline that realizes informative touches in practice, combining reachability-aware target selection, motion planning, compliant execution, and tactile servoing for stable contact data acquisition. \textbf{Third}, we demonstrate in simulation and on a real Franka Panda robot that integrating the full spectrum of contact-derived signals reduces the number of action cycles required for correct classification and improves pose accuracy and stability compared to baselines that use only subsets of these sensing modalities.



\section{Related work}

Our work is closely related to recent works that employ tactile sensing in Bayesian-based object recognition, pose estimation, and active tactile exploration.
Static tactile object recognition was tackled in~\cite{Schmitz_2014}, where they recognize grasped object with an almost fully sensorized hand using deep learning technique. Other approaches address the problem on static conditions focusing on small objects~\cite{patel2021diggerfingergelsighttactile, Karamipour_2023}. Previous work approached the object recognition problem based on material properties. In this context,~\cite{kaboliTactilebasedActiveObject2019,kaboliTactileBasedFrameworkActive2017,xuTactileIdentificationObjects2013, regoliControlledTactileExploration2017} relied on properties such as stiffness, texture, thermal conductivity, and softness to distinguish between different objects, while our method focuses on object recognition with geometric information.
Pose estimation from single touch was tackled in~\cite{bauzaTactileObjectPose2020, Bauza_2023}, where a perception model was trained in simulation to estimate a probability distribution over possible object poses given tactile data. In~\cite{caddeoCollisionawareInhand6D2023} the authors leverage tactile sensing to infer poses compatible with the surface touched. Because tactile observations are inherently local and sparse, a single touch may not sufficiently reduce uncertainty over object class and 6D pose, motivating belief updates from sequential tactile measurements.
Bayesian filtering was employed in~\cite{siposSimultaneousContactLocation2022} to fuse proprioception and touch and estimate both the
location of contacts and the in-hand object pose. Some researchers used methods more in line with our work with touch-based pose estimation of objects. Koval et al.~\cite{kovalPoseEstimationPlanar2015} proposed the manifold particle filter by adaptively sampling particles that reside on the contact manifold to increase sampling efficiency.  In~\cite{petrovskayaGlobalLocalizationObjects2011} the authors proposed the Scaling Series algorithm, combining the Bayesian Monte Carlo technique coupled with annealing, to refine the posterior distribution of the 6-degree-of-freedom (DOF) pose through multiple stages with a small number of particles at each stage. Vezzani et al.~\cite{vezzaniMemoryUnscentedParticle2017} proposed using a memory unscented Kalman filter (MUKF) to cover the 6-DOF pose space with a small number of particles, where each particle is treated as a Gaussian distribution instead of a discrete sample. 
Although the methods are quite efficient, the works by~\cite{ kovalPoseEstimationPlanar2015, petrovskayaGlobalLocalizationObjects2011, vezzaniMemoryUnscentedParticle2017} were confined to pose estimation. 
As they limited their scope to the pose estimation of a known object, multi-class classification was not addressed in their work. Vezzani et al.~\cite{vezzaniNovelBayesianFiltering2016} extended their work on pose estimation~\cite{vezzaniMemoryUnscentedParticle2017} to address object recognition and pose estimation simultaneously by applying the localization scheme to multiple object classes and selecting the object class with the smallest localization error.~\cite{zheng2024bayesianframeworkactiveobject} proposes a unified Bayesian framework for active tactile object recognition, pose estimation, and shape reconstruction using object priors under idealized assumptions. As a largely theoretical study, it models the tactile sensor as a point contact and does not exploit the multimodal signals available in our setting, including GelSight observations, wrist force/torque measurements, and motion-derived free-space information.

Active data acquisition is crucial to our approach and strongly affects recognition and pose estimation performance. Similarly to our approach, an active tactile perception algorithm was devised in~\cite{Xiao_2022}, leveraging contour tracing to recognize the object. In~\cite{10611667} the authors introduce an active exploration approach based on reinforcement learning to precisely reconstruct the shape of objects using vision-based tactile sensors. In~\cite{bonzini2025robotic} the authors devise a novel exploration strategy to simultaneously detect symmetries in a 3-D object and use this information to enhance shape estimation, leveraging the Gaussian Process (GP). Similarly, Driess~\cite{driessActiveLearningQuery,driess_active_2019} et. al. used the GP posterior variance and differential entropy to guide the active exploration of robot fingers on surface of an object for shape reconstruction. The authors of~\cite{Tandem3d_2023} expand the 2D problem of~\cite{Tandem_2022} with a co-training framework for exploration and decision making to 3D object recognition with tactile signals. Instead of adopting data-hungry reinforcement-learning approaches or GP-based planners with substantial computational overhead, we propose a lightweight planning and control stack with a simple but effective target-point selection strategy for active data acquisition, tailored to the physical constraints of our hardware and workspace.

\section{Method}\label{sec:method}
The goal of this paper is to incorporate the tactile information made available by the vision-based tactile sensor, force torque sensor and robot proprioception to achieve fast active object recognition and pose estimation with a real robot setup. On top of that, a motion planning and control framework is developed and integrated with the Bayesian framework to perform active data acquisition using a Franka Emika panda robot arm with a force torque sensor and a GelSight tactile sensor mounted as the end effector.

\begin{figure*}[!t]
  \centering
  \centerline{\includegraphics[width = \linewidth]{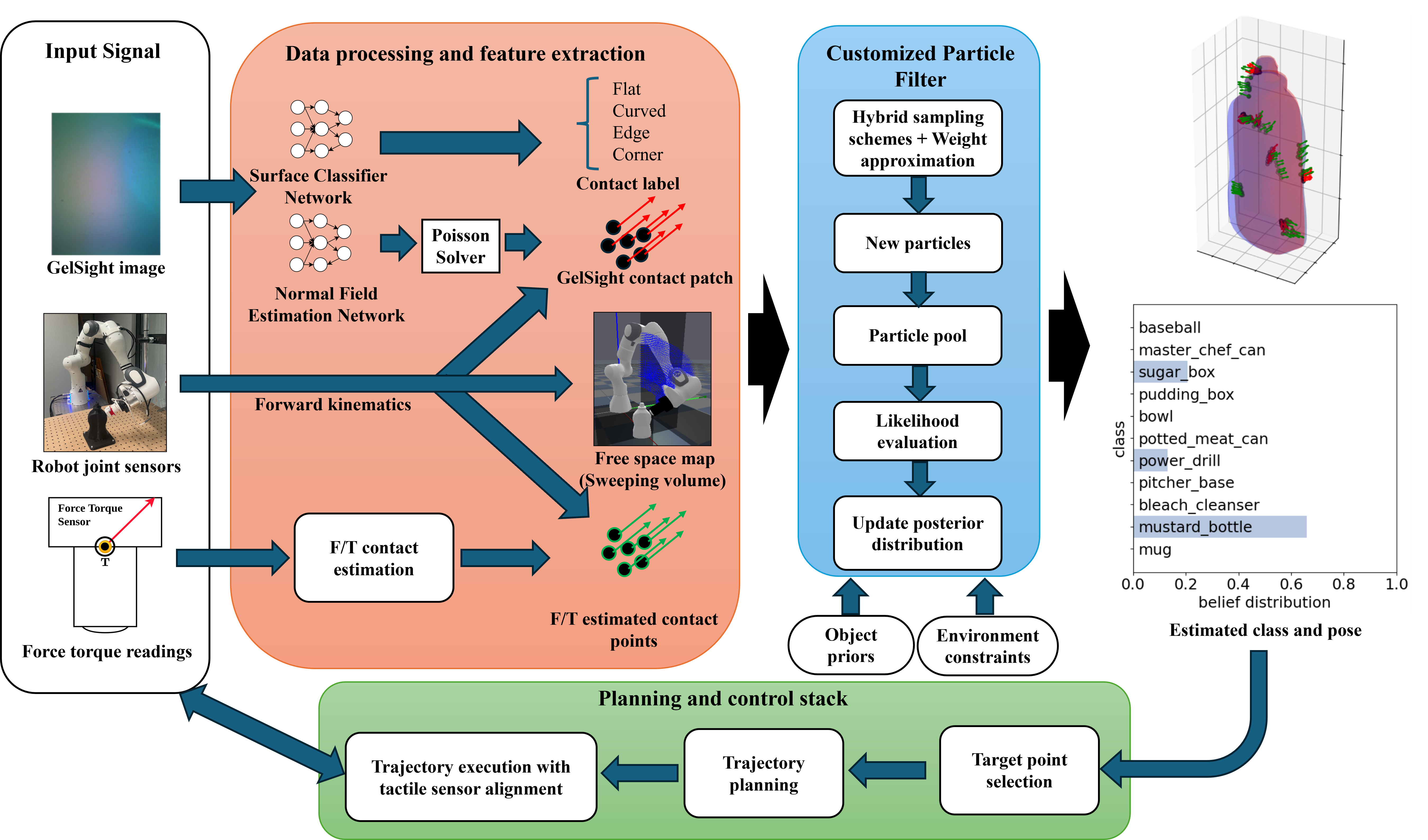}}
  \caption{Overview of the proposed system. The system processes measurements from the GelSight sensor, the force/torque sensor, and robot proprioception to extract contact points and contact-type labels, and to build a free-space map. These observations are then fed to a customized particle filter, which performs Bayesian inference over object class and 6D pose. The particle filter proposes new particles conditioned on the incoming observations, approximates their weights, and mixes them with the existing particle set for subsequent belief updates, yielding both the MAP estimate and the posterior distribution. The MAP estimate is used to select the next target point and approach direction for data collection. Given the target point and approach direction, our planning and control stack generates and executes a trajectory with compliance, and aligns the GelSight sensor with the contact surface once contact is detected. Each module shown in the figure is described in detail in the Section~\ref{sec:method}.
}
  \label{fig:overview}
\end{figure*}

This section provides a detailed description of all the components shown in Fig.~\ref{fig:overview}. Firstly, a description of the data extraction methods for the GelSight Sensor, force torque sensor and robot motion is given. Then the derivation of the Bayesian probabilistic model used to fuse the extracted data for object class and pose inference is presented. Lastly, implementation details on the target point selection scheme and the developed motion planning and control framework for active data collection are given.
The world frame in this study is defined as the base frame of the robot arm, the estimated object pose are expressed in the world frame.

\subsection{Data processing and feature extraction}
\subsubsection{Data extraction from GelSight Mini}
To refine joint estimation of object class and pose, we use contact information from GelSight tactile sensors, focusing on the local geometric features of the surface in contact. From a GelSight Mini sensor, a tactile image is obtained based on the force distribution at the contact patch, from their tactile image we reconstruct surface normals and depth using the GelSight pipeline~\cite{Yuan2017}, and form an oriented point cloud; in simulation we use TACTO-generated depth~\cite{wangTACTOFastFlexible2022}. For each contact patch, we retain points deeper than 0.5 mm, downsample the points to 10 points per patch for efficiency, and transform them to the world frame via forward kinematics before using them for Bayesian inference. In parallel, a contact classifier~\cite{10851808} predicts one of four local contact types (flat, curved, edge, corner). The oriented points and contact label are then used as observations to infer object class and pose.

\subsubsection{Contact Estimation with Force Torque Sensor}
Precise GelSight alignment is not always feasible due to kinematic and collision constraints. We therefore introduce a force/torque-based contact estimation module that uses wrench measurements and the known end-effector geometry to infer the contact point during collisions. Unlike with the Gelsight sensor, this method assumes a single point contact with no free moment, enabling real-time estimation. While accuracy degrades when multiple simultaneous contacts occur (e.g., in concave regions), the estimates remain informative.

\begin{figure}[thpb]
  \centering
  \includegraphics[width=\linewidth]{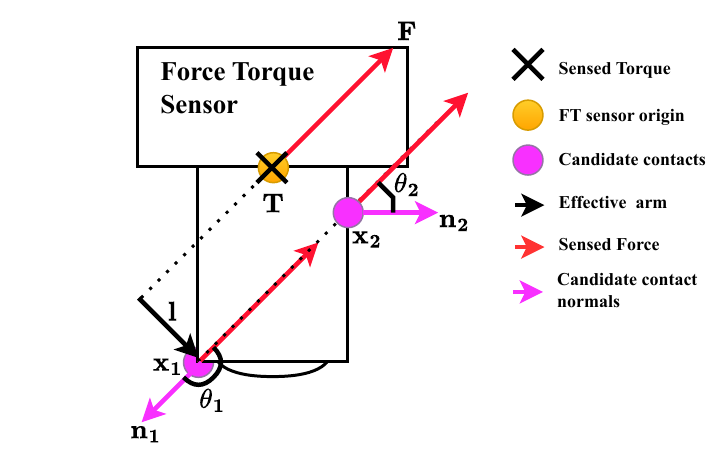}
  \caption{Example of force/torque-based contact localization. The cross indicates the outward direction of the measured torque $\mathbf{T}$. From the measured wrench $(\mathbf{F},\mathbf{T})$, two candidate contact points $\mathbf{x}_1$ and $\mathbf{x}_2$ are computed. The estimate $\mathbf{x}^*$ is chosen as $\mathbf{x}_1$ since $\angle(\mathbf{n}_1,\mathbf{F})>135^\circ$ while $\angle(\mathbf{n}_2,\mathbf{F})<135^\circ$.
}
  \label{fig:FT_contact_estimation}
\end{figure}

A graphical example of contact location estimation based on force torque readings is given in Fig.~\ref{fig:FT_contact_estimation}. To estimate the contact location given the force reading $\mathbf{F} = [F_x,F_y,F_z]$ and torque reading $\mathbf{T} = [T_x,T_y,T_z]$ from the force torque sensor, we calculate the effective arm vector $\mathbf{l_0}$ by 
\begin{equation}
  \mathbf{l_0} = \frac{\mathbf{F} \times \mathbf{T}}{||\mathbf{F}||^2} 
\end{equation}
For each vertex $\mathbf{x}$ on the end effector, we calculate its projection on the effective arm vector direction, 
\begin{equation}
   \mathbf{l}(\mathbf{x}) = \frac{\mathbf{x}\cdot\mathbf{l_0}}{||\mathbf{l_0}||^2}\mathbf{l_0}
\end{equation}
The distance between $\mathbf{l}(\mathbf{x})$ and $\mathbf{l}_0$, i.e., $\|\mathbf{l}(\mathbf{x})-\mathbf{l}_0\|$, measures how well a candidate vertex $\mathbf{x}$ explains the measured wrench. However, minimizing $\|\mathbf{l}(\mathbf{x})-\mathbf{l}_0\|$ alone typically yields two candidate contact locations on a convex body with different surface normals. We therefore incorporate normal--force consistency to disambiguate the solution.

Assuming the friction coefficient $\mu$ between the end effector and the object is less than 1, the reaction force direction $-\mathbf{F}$ must lie within the friction cone. Using the conservative bound $\mu<1 $
implies the angle between $-\mathbf{F}$ and the contact normal $\mathbf{n}(\mathbf{x})$ should be smaller than $45^\circ$. We estimate the contact location as
\begin{equation}
\mathbf{x}^*=\arg\min_{\mathbf{x}\in\mathbf{V}_e}
\|\mathbf{l}(\mathbf{x})-\mathbf{l}_0\|
+\lambda\,\mathds{1}\!\left[
\frac{\mathbf{n}(\mathbf{x})\cdot(-\mathbf{F})}{\|\mathbf{F}\|} < \cos\!\left(\frac{\pi}{4}\right)
\right],
\end{equation}
where $\mathds{1}[\cdot]$ is an indicator function. The penalty weight $\lambda$ is chosen such that violating the friction-cone constraint is penalized equivalently to a $1\,\mathrm{m}$ displacement in the effective moment arm.

The estimated contact location $\mathbf{x}^*$ is then transformed from the F/T sensor frame to the world frame using forward kinematics and passed to the Bayesian inference framework. Contact estimation is performed only when $\|\mathbf{F}\|>2\,\mathrm{N}$ to ensure sufficient signal-to-noise ratio. To reduce variance in the estimated contact position, we apply a sliding-window average over the 10 most recent predictions.

\subsubsection{Free space carving with robot motion}
As the robot moves, its end-effector sweeps out a volume that certifies free space. This information could be leveraged to rule out object classes and pose hypotheses that violate the non-penetration constraints with the free space.
To store and update the free space carved out by the robot, we maintain a 3D voxel grid map of the workspace. The workspace is assumed to be 1m x 1m x 1m centered at (0,0.5,0.5) in the base frame of the robot. It is discretized into a $(200 \times 200 \times 200)$ grid map, each grid is of size $(5\text{mm} \times 5\text{mm} \times 5\text{mm})$. 
The map is initialized to zero and updated at 60 Hz using the current end-effector volume. Let $\mathbf{V}_e$ denote a set of vertices sampled on the last two links of the robot arm and the end-effector geometry; we mark all voxels containing any vertex as free 
\begin{equation}
   m^r(\mathbf{x})= 1, \forall \mathbf{x} \in \mathbf{V}_e.
\end{equation}
Here $m^r(\mathbf{x})$ indexes the voxel containing $\mathbf{x}$, Any voxel with value greater than $0$ is treated as free space and used to reject object class/pose hypotheses that would intersect it.

\subsection{Bayesian joint estimation of object class and object pose}
We formulate the joint estimation of the object class and object pose as a Bayesian inference problem. \\
\textbf{Latent variables:} object class $c$ and object pose $\mathbf{p}$, concatenated into a vector $\mathbf{z} \leftarrow \left [ c, \mathbf{p} \right ]$.\\ 
\textbf{Object priors:} We represent each object in the form of a signed distance field (SDF). Given an object class $c$ and object pose $\mathbf{p}$, the predicted Euclidean signed distance value $\widehat{d}_s$ from a point $\mathbf{x}$ to the object surface is given by its signed distance function
\begin{equation}
    \widehat{d}_s := f(\mathbf{x}, \mathbf{z}),
\end{equation}
\begin{equation}
    \begin{cases} 
      \widehat{d}_s=0 &  \text{on surface}\\
      \widehat{d}_s>0 & \text{outside object}\\
      \widehat{d}_s<0 & \text{inside object }
   \end{cases}
\end{equation}
The predicted surface normal $\mathbf{\widehat{n}}$ at a contact point $\mathbf{x}$ can be obtained through differentiation, 
\begin{equation}
    \mathbf{\widehat{n}} = \nabla_{\mathbf{x}}{f}(\mathbf{x}, \mathbf{z}).
\end{equation}
\textbf{Known environment constraints:} In this study, experiments are performed on a table in front of the robot arm, which serves as the floor of the environment. We assume the table height is known. Under gravity, the object should rest on the table; thus it should neither float above nor penetrate the tabletop. We denote these tabletop constraints by $\mathbf{E}$ and write $p(\mathbf{E}\mid\mathbf{z})$ for the likelihood that $\mathbf{E}$ is satisfied given $\mathbf{z}$. These constraints are enforced at every time step.\\
\textbf{Input:} At time step $t$, the robot may receive oriented contact points from the GelSight sensor, $(\mathbf{X}^g_t,\mathbf{N}^g_t)$, oriented contact points from the force/torque sensor, $(\mathbf{X}^f_t,\mathbf{N}^f_t)$, and the free-space map $\mathbf{M}^r_t$, where superscripts $g$, $f$, and $r$ denote GelSight, FT sensing, and robot motion, respectively.\\
We represent each oriented point as a single observation
$\mathbf{d}_{i,t} \triangleq \big[\mathbf{x}_{i,t},\,\mathbf{n}_{i,t}\big]$ with $\mathbf{x}_{i,t}\in\mathbf{X}_t$ and $\mathbf{n}_{i,t}\in\mathbf{N}_t$, and collect all observations in $\mathbf{D}_t$. Superscripts are used to distinguish sources (e.g., $g$ vs.\ $f$); when omitted, the notation refers to the union of all contacts.\\
\textbf{Assumption:} The object remains stationary during tactile exploration.\\ 
\textbf{Goal:} Estimate the posterior distribution of the latent variables $p(\mathbf{z} | \mathbf{M^{r}}_{1:T},\mathbf{D}^{g}_{1:T},\mathbf{D}^{f}_{1:T}, \mathbf{E}_{1:T})$ given the observations and the environmental constraints from the time step $1$ to $T$. Assuming conditional independence in observations across modalities and different time steps when conditioning on $\mathbf{z}$, based on Bayes' theorem, we obtain the following equations,
\begin{align}
\begin{aligned}
      &p(\mathbf{z} | \mathbf{M^{r}}_{1:T},\mathbf{D}^{g}_{1:T},\mathbf{D}^{f}_{1:T}, \mathbf{E}_{1:T}) \\
      &= \frac{p(\mathbf{z}, \mathbf{M^{r}}_{1:T},\mathbf{D}^{g}_{1:T}, \mathbf{D}^{f}_{1:T}, \mathbf{E}_{1:T})}{\sum_{\mathbf{z}}p(\mathbf{z}, \mathbf{M^{r}}_{1:T},\mathbf{D}^{g}_{1:T}, \mathbf{D}^{f}_{1:T}, \mathbf{E}_{1:T})}\\
      &\propto p(\mathbf{z}, \mathbf{M^{r}}_{1:T},\mathbf{D}^{g}_{1:T}, \mathbf{D}^{f}_{1:T}, \mathbf{E}_{1:T})\\
      &= p(\mathbf{z})p(\mathbf{M}^{r}_{1:T}|\mathbf{z})p(\mathbf{D}^{g}_{1:T}|\mathbf{z})p(\mathbf{D}^{f}_{1:T}|\mathbf{z})p(\mathbf{E}_{1:T}|\mathbf{z})\\
      &= p(\mathbf{z})p(\mathbf{M}^{r}_{1:T-1}|\mathbf{z})p(\mathbf{D}^{g}_{1:T-1}|\mathbf{z})p(\mathbf{D}^{f}_{1:T-1}|\mathbf{z})p(\mathbf{E}_{1:T-1}|\mathbf{z})\\
      &\times p(\mathbf{M}^{r}_{T}|\mathbf{z})p(\mathbf{D}^{g}_{T}|\mathbf{z})p(\mathbf{D}^{f}_{T}|\mathbf{z})p(\mathbf{E}_{T}|\mathbf{z})\\
      &= \underbrace{p(\mathbf{z}, \mathbf{M}^{r}_{1:T-1},\mathbf{D}^{g}_{1:T-1}, \mathbf{D}^{f}_{1:T-1},\mathbf{E}_{1:T-1})} _{\text{joint distribution in the previous time steps}}\times\\
      & \underbrace{p(\mathbf{M}_{T}|\mathbf{z})p(\mathbf{D}^{g}_{T}|\mathbf{z})p(\mathbf{D}^{f}_{T}|\mathbf{z})p(\mathbf{E}_{T}|\mathbf{z})}_{\text{observational likelihood in the current time step}}  
\end{aligned}
\label{eqn:posterior_all_time}
\end{align}
In this study, $p(\mathbf{z})$ is an uniform distribution.
Since $\mathbf{z}=(c,\mathbf{p})$ comprises a discrete class variable $c$ and a continuous pose $\mathbf{p}$, marginalizing over $\mathbf{z}$ amounts to summing over $c$ and integrating over $\mathbf{p}$. With a slight abuse of notation, we write $\sum_{\mathbf{z}}$ to denote this sum--integral operation. Equation~\eqref{eqn:posterior_all_time} then yields the recursive posterior update. Assuming conditional independence of oriented contact points within a time step given $\mathbf{z}$, the likelihoods $p(\mathbf{D}^{g}_{t}\mid \mathbf{z})$ and $p(\mathbf{D}^{f}_{t}\mid \mathbf{z})$ further decompose as
\begin{align}
   p(\mathbf{D}^{g}_{t}|\mathbf{z}) =\underset{i=1}{\prod^{N_g}} p(\mathbf{d}^{g}_{i,t}|\mathbf{z}),\\
   p(\mathbf{D}^{f}_{t}|\mathbf{z}) =\underset{i=1}{\prod^{N_f}} p(\mathbf{d}^{f}_{i,t}|\mathbf{z}).
   \label{eqn:decomposition}
\end{align}
\subsection{Likelihood function for oriented contact points}
To derive the likelihood function for oriented contact points, we treat the difference $\Delta d$ between the signed distance value $d_s$ and the predicted signed distance value $\widehat{d}_s$, and the angle $\theta$ between the normal vector $\mathbf{n}$ and the predicted normal vector $\mathbf{\widehat{n}}$ as two independent random Gaussian variables, where $\Delta d \sim \mathcal{N}(0,\sigma_d)$ and $\theta \sim \mathcal{N}(0,\sigma_{\theta})$. For contact points $d_s=0$, the measurement likelihood function can be written as 
\begin{align}
    \theta(\mathbf{x,\mathbf{z}})& = \arccos(\mathbf{\widehat{n}}(\mathbf{x,\mathbf{z}})\cdot\mathbf{n}),\\
    p(\mathbf{d}|\mathbf{z})  
       &=\frac{1}{Z} \exp\left [ -\frac{1}{2}\left(\frac{{\widehat{d}_s}^{2}(\mathbf{x,\mathbf{z}})}{\sigma_d^{2}}+\frac{\theta^2(\mathbf{x,\mathbf{z}})}{\sigma_\theta^2}\right)\right ],
\label{eqn:pf_likelihood}
\end{align}

where $Z$ is the normalization constant for the Gaussian likelihood function.
Note $\sigma_d$ and $\sigma_\theta$ for GelSight contact points and FT contact points take different values, where $\sigma_{d}^{f}=2\sigma_{d}^{g}$ and $\sigma_{\theta}^{f}=2\sigma_{\theta}^{g}$.

\subsection{Contact label for GelSight contact points}
In addition, each GelSight contact patch $\mathbf{D}^{g}_{t}$ comes with a contact type label. If the detected contact type label is an edge or a corner, the likelihood calculation would not take into account the normal vectors in that patch for the likelihood calculation. In this case, for each oriented contact observation
$\mathbf d^g_{i,t} = [\mathbf x^g_{i,t}, \mathbf n^g_{i,t}] \in \mathbf D^g_t$, we use
\begin{equation}
p(\mathbf d^g_{i,t}\mid \mathbf z)
= \frac{1}{Z}\exp\!\left(
-\frac{{\widehat d_s}^2(\mathbf x^g_{i,t},\mathbf z)}{2{\sigma^g_d}^{2}}
\right).
\end{equation}
The reasoning behind this choice is that at sharp edges and corners, the gradient on the signed distance function is ill-defined.


\subsection{Likelihood function for non-penetration constraint with the free space carved out by robot motion}
To evaluate the extent of violation of the non-penetration constraint for a hypothesis $\mathbf{z}$, we represent each hypothesis $\mathbf{z}$ with 200 feature points $\mathbf{V}(\mathbf{z})$ on the estimated object surface, which is obtained by transforming 200 precomputed feature points of the hypothesized class into the hypothesized pose. Then we lookup the corresponding values of the transformed points on the free space grid map, and check the number of feature points that overlap with the free space grids, denoted by $n$, where
\begin{equation}
    n = \sum_{\mathbf{v} \in \mathbf{V}(\mathbf{z})} m^r(\mathbf{v}),
\end{equation}
$m^r$ denotes the free space map lookup function.

The observational likelihood function for observing the free space map $\mathbf{M^r}$ given $\mathbf{z}$ is defined as
\begin{equation}
    p(\mathbf{M^r}| \mathbf{z} ) = \frac{1}{Z_m}\exp(-\frac{1}{2{\sigma^g_d}^2} \cdot (1)^2 \cdot n),
    \label{eqn:free_space_likelihood}
\end{equation}
where $Z_m$ is the normalization constant.

In this study, the free-space map is enforced as a semi-hard constraint that strongly penalizes particles violating non-penetration while avoiding zero weights even when all particles violate the constraint. Each violation reduces the likelihood by an amount equivalent to observing a contact point located 1\,m from the estimated object surface. To avoid over-penalizing minor mismatches, we update the free-space map using a slightly shrunken end-effector volume (1\,cm margin), so particles that marginally intersect the true end effector are not penalized.
A key advantage of this scheme is that the per-hypothesis evaluation cost remains constant as the free-space map grows in complexity.

\subsection{Likelihood function for environment constraints}
To capture the tabletop constraint, we denote the height of the table as $h_t$ and define the following likelihood function to penalize object hypotheses whose lowest surface point is more than $2 cm$ above or below the table plane,
\begin{equation}
    p(\mathbf{E}| \mathbf{z} ) = \frac{1}{Z_e}\exp\left(-\frac{\mathds{1}(v_z^*>h_t+0.02)+\mathds{1}(v_z^*<h_t-0.02)}{2{\sigma^g_d}^2}\right),
    \label{eqn:environment_likelihood}
\end{equation}
\begin{equation}
    v_z^* = \min_{\mathbf{v} \in \mathbf{V}(\mathbf{z})} v_z,
\end{equation}
where $\mathds{1}(\cdot)$ is the indicator function and $Z_e$ is the normalization constant. A violation of the constraint reduces the
likelihood by an amount equivalent to observing a contact
point located 1 m from the estimated object surface. Similarly to the free-space likelihood function, this formulation avoids
over-penalizing minor mismatches.
\subsection{Customized particle filter for object recognition and pose estimation}
\begin{figure}[thpb]
  \centering
  \includegraphics[width=\linewidth]{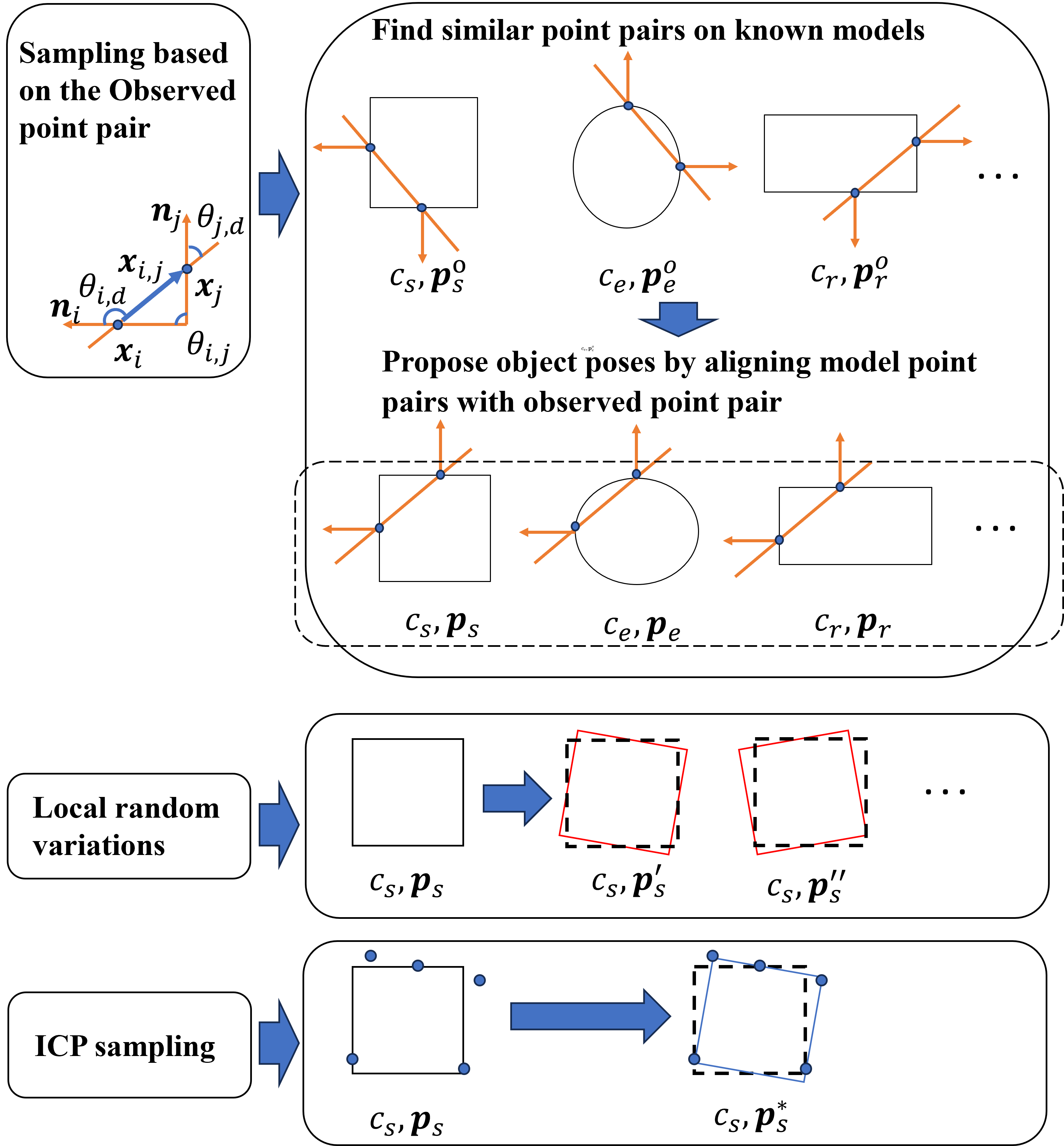}
  \caption{Example of the three particle sampling schemes. \emph{Point-pair-based sampling:} for each observed contact point pair $(\mathbf{x}_i,\mathbf{n}_i)$ and $(\mathbf{x}_j,\mathbf{n}_j)$, we compute point-pair features $\|\mathbf{x}_{i,j}\|_2$, $\theta_{i,j}$, $\theta_{i,d}$, $\theta_{j,d}$, and labels $l_i,l_j$, where $\mathbf{x}_{i,j}=\mathbf{x}_j-\mathbf{x}_i$. Here $\theta_{i,j}$ is the angle between $\mathbf{n}_i$ and $\mathbf{n}_j$, while $\theta_{i,d}$ and $\theta_{j,d}$ are the angles between $\mathbf{n}_i$ and $\mathbf{x}_{i,j}$, and between $\mathbf{n}_j$ and $\mathbf{x}_{i,j}$, respectively. Model point pairs with similar features are retrieved from the database; aligning a model pair to the observed pair yields candidate class/pose hypotheses. \emph{Local random variations:} additional particles are generated by applying small perturbations to the poses of the top-weighted particles (e.g., $\mathbf{p}_s \rightarrow \mathbf{p}'_s,\mathbf{p}''_s$). \emph{Iterative closest point(ICP) sampling:} ICP is applied to refine the poses of top particles using the accumulated contact points, producing new hypotheses that better fit the observations (e.g., $\mathbf{p}_s \rightarrow \mathbf{p}^*_s$). In the illustration, $c_s,c_e,c_r$ denote three object classes (square, ellipse, and rectangle), with canonical poses $\mathbf{p}^o_s,\mathbf{p}^o_e,\mathbf{p}^o_r$ and aligned poses $\mathbf{p}_s,\mathbf{p}_e,\mathbf{p}_r$.
 }
  \label{fig:propose_particles}
\end{figure}
Combining \eqref{eqn:posterior_all_time}--\eqref{eqn:environment_likelihood} yields a recursive Bayesian update for
$p(\mathbf{z}\mid \mathbf{M}^{r}_{1:t},\mathbf{D}^{g}_{1:t},\mathbf{D}^{f}_{1:t},\mathbf{E}_{1:t})$. In practice, the normalization over $\mathbf{z}$ is intractable, so we approximate the posterior with a particle filter using a finite number of
samples based on importance sampling:
\begin{equation}
    p(\mathbf{z} |\mathbf{M}^{{r}}_{1:t}, \mathbf{D}^{{g}}_{1:t},\mathbf{D}^{{f}}_{1:t},\mathbf{E}_{1:t}) = \sum_{j=1}^{N} \overline{w}_{j,t} \; \delta(\mathbf{z}-\mathbf{z}_{j,t}),
    \label{eqn:importance_sampling}
\end{equation}
where $\delta(\cdot)$ is the Dirac delta and $\{\mathbf{z}_{j,t},\overline{w}_{j,t}\}_{j=1}^{N}$ are weighted particles. Although the subscript $t$ typically denotes a time-varying state~\cite{elfringParticleFiltersHandsOn2021}, we assume that the object is stationary; thus $\mathbf{z}_{j,t+1}=\mathbf{z}_{j,t}$ and only the weights are updated over time. The resulting recursive update for $\overline{w}_{j,t}$ is derived as follows:
\begin{equation}
\begin{aligned}
        \overline{w}_{j,t} &= p(\mathbf{z}_{j,t} |\mathbf{M}^{{r}}_{1:t}, \mathbf{D}^{{g}}_{1:t},\mathbf{D}^{{f}}_{1:t},\mathbf{E}_{1:t}) \\
        & \propto w_{j,t}=p(\mathbf{z}_{j,t}, \mathbf{M}^{{r}}_{1:t},\mathbf{D}^{{g}}_{1:t},\mathbf{D}^{{f}}_{1:t},\mathbf{E}_{1:t})\\
        & = p(\mathbf{z}_{j,t},\mathbf{M}^{{r}}_{1:t}, \mathbf{D}^{{g}}_{1:t-1},\mathbf{D}^{{f}}_{1:t-1},\mathbf{E}_{1:t-1})\; p( \mathbf{M}^{{r}}_{t}, \mathbf{D}^{{g}}_{t},\mathbf{D}^{{f}}_{t},\mathbf{E}_{t} | \mathbf{z}_{j,t})\\
        &= \underbrace{p(\mathbf{z}_{j,t}, \mathbf{M}^{{r}}_{1:t-1}, \mathbf{D}^{{g}}_{1:t-1},\mathbf{D}^{{f}}_{1:t-1},\mathbf{E}_{1:t-1})\;}_{{w}_{j,t-1}}\\
        &\times \underbrace{p(\mathbf{M}^{{r}}_{t}| \mathbf{z}_{j,t})p(\mathbf{D}^{{g}}_{t}| \mathbf{z}_{j,t}) p(\mathbf{D}^{{f}}_{t}| \mathbf{z}_{j,t}) p(\mathbf{E}_{t}| \mathbf{z}_{j,t})}_\text{observational likelihood in the current step}.
\end{aligned}
\label{eqn:importance_weight}
\end{equation}
In \eqref{eqn:importance_weight}, the unnormalized weight $w_{j,t}$ is calculated, to get $\overline{w}_{j,t}$, we divide it by the sum of all $w_{j,t}$,
\begin{equation}
    \overline{w}_{j,t} = \frac{w_{j,t}}{\sum_{j=1}^{N} w_{j,t}}.
    \label{eqn:importance_weight_normalized}
\end{equation}

Approximating the joint posterior over 11 object classes and a 6D pose requires many particles, especially early in inference when uncertainty is high. To keep the particle set tractable, we adapt the point-pair feature sampling scheme and weight approximation scheme of~\cite{zheng2024bayesianframeworkactiveobject} to progressively propose new particles at each time step conditioned on the latest observations (contact points, normals), rather than uniformly covering the full latent space. Figure~\ref{fig:propose_particles} summarizes the sampling scheme. At each time step, we downsample the latest patch of contact points to pair with downsampled previously observed contact points. For each point pair, we calculate the distance (rounded to the nearest 0.01 cm) between the two points, the angles (rounded to the nearest multiples of 12 degrees) between their respective normals and the displacement vectors as its features. Potential point pair correspondences on known models can be found by matching the point-pair features of each observed point pair with the precomputed feature values for the point pairs on known models (stored in a hash table for fast lookup). Aligning the corresponding point pairs on known models with observed point pairs yields candidate class/pose hypotheses.

A limitation of point-pair sampling scheme is that it relies on a discretized surface (vertices), which can introduce small residual pose offsets even in the noise-free case. To mitigate this limitation, we supplement point-pair sampling with two additional proposal mechanisms. First, we refine the top 10 MAP particles using iterative closest point (ICP) against the observed contact points, and insert the refined poses as new particles. Since ICP is unreliable with sparse contacts, we activate it only after collecting more than four GelSight contact patches. Before ICP is enabled, we instead jitter the top 10 MAP particles by sampling independent translation perturbations in each axis from $U[-0.01,0.01]$\,m and rotation perturbations in each Euler angle from $U[-10,10]^\circ$, generating 12 variants per particle (120 total). The same weight approximation scheme is applied to all newly proposed particles at a given time step, independent of the proposal strategy.
For new particles $\mathbf{z}^{'}_{k,t}$, $k = 1,2,3,\dots$ proposed at time step $t$, we approximate the ratio of the weights $w^{'}_{k,t}$ among the newly proposed particles by evaluating them on a subset of observed contact points $\mathbf{D}^{g}_{S}$ and $\mathbf{D}^{f}_{S}$, specifically by the following equation,
\begin{equation}
    \frac{w^{'}_{k,t}}{w^{'}_{*,t}} = \frac{p(\mathbf{D}^{g}_{S}, \mathbf{D}^{f}_{S}| \mathbf{z}^{'}_{k,t})}{p(\mathbf{D}^{g}_{S}, \mathbf{D}^{f}_{S}| \mathbf{z}^{'}_{*,t})},
    \label{eqn:relative_weight_1}
\end{equation}
where
\begin{equation}
    \mathbf{z}^{'}_{*,t} = \argmax_{\mathbf{z}^{'}_{k,t}} p(\mathbf{D}^{g}_{S}, \mathbf{D}^{f}_{S}| \mathbf{z}^{'}_{k,t}).
\end{equation}
$\mathbf{D}^{g}_{S}$ and $\mathbf{D}^{f}_{S}$ are down-sampled from existing contact points using Poisson Disk Sampling, such that they do not exceed a predefine number of points limit (e.g., 100 points) to keep the computational cost tractable. 
Then the ratio between the maximal weight $w^{'}_{*,t}$ among the newly proposed particles and the MAP weight $\overline{w}^{*}_{t}$ by evaluating them on all observed contact points $\mathbf{D}^{g}_{1:t}$, $\mathbf{D}^{f}_{1:t}$ by
\begin{equation}
    \frac{w^{'}_{*,t}}{\overline{w}^{*}_{t}} = \frac{p(\mathbf{D}^{g}_{1:t},\mathbf{D}^{f}_{1:t}| \mathbf{z}^{'}_{*,t})}{p(\mathbf{D}^{g}_{1:t},\mathbf{D}^{f}_{1:t}| \mathbf{z}^{*}_{t})},
    \label{eqn:relative_weight_2}
\end{equation}
where 
\begin{equation}
    \mathbf{z}^{*}_{t} = \argmax_{\mathbf{z}_{j,t}}\overline{w}_{j,t}.
\end{equation}
Combining \eqref{eqn:relative_weight_1}-\eqref{eqn:relative_weight_2} yields,
\begin{equation}
    w^{'}_{k,t} = \frac{p(\mathbf{D}^{g}_{S},\mathbf{D}^{f}_{S}| \mathbf{z}^{'}_{k,t})p(\mathbf{D}^{g}_{1:t},\mathbf{D}^{f}_{1:t}| \mathbf{z}^{'}_{*,t})}{p(\mathbf{D}^{g}_{S},\mathbf{D}^{f}_{S}| \mathbf{z}^{'}_{*,t})p(\mathbf{D}^{g}_{1:t},\mathbf{D}^{f}_{1:t}| \mathbf{z}^{*}_{t})}\overline{w}^{*}
    \label{eqn:new_weights}
\end{equation}
This approximation scheme keeps the per-step likelihood evaluation tractable by scoring most newly proposed particles on only a subset of the observed contact points. Only the best-performing new particle on this subset, denoted $\mathbf{z}'_{j,t}$, is then evaluated on the full contact set, ensuring that a new particle can overtake the current MAP estimation only if it fits all observations better. The top-$K$ new particles are retained and merged into the particle pool.

To further reduce computation, the free-space map and environment constraints are omitted during this approximation stage and are only evaluated for all particles after the new particles have been added to the particle pool.


\subsection{Target point selection}
To actively acquire data, we select the next target point via the following criterion:
\begin{equation}
\overline{\mathbf{x}}_{t+1}
=\argmax_{\mathbf{x}_m\in\mathbf{M}^*}\;
\min_{\mathbf{x}_c\in \mathbf{X}_c\cup \mathbf{X}_u}
\|\mathbf{x}_m-\mathbf{x}_c\|_2,
\label{eqn:exploration_DHD}
\end{equation}
where $\mathbf{M}^*$ is the set of candidate vertices on the MAP shape, $\mathbf{X}_c$ are observed contact points, and $\mathbf{X}_u$ are previously selected but unreachable targets. This favors points far from both explored and unreachable regions.

Since experiments are performed on a table, we restrict candidates to vertices above height $h$ whose normals do not point downward:
\begin{equation}
\mathbf{M}^{*}=\left\{\mathbf{x}\in\mathbf{M}\ \middle|\ x_z>h,\ \mathbf{n}(\mathbf{x})\cdot[0,0,1]^T\ge 0\right\}.
\label{eqn:physical_contraints}
\end{equation}
The target point $\overline{\mathbf{x}}_{t+1}$ with its corresponding normal vector $\overline{\mathbf{n}}_{t+1}$ is then passed to a motion planner for trajectory planning.
If no feasible trajectory to reach $\overline{\mathbf{x}}_{t+1}$ is found, $\overline{\mathbf{x}}_{t+1}$ will be added to $\mathbf{X}_u$, and a new target point will be proposed using \eqref{eqn:exploration_DHD} with updated $\mathbf{X}_u$.


\subsection{Motion planner and Control Scheme for Active Data Acquisition}
Given a proposed target point with normal direction, the robot must plan and carry out motions to establish contacts with the object surface and collect free-space information for Bayesian inference. Due to high uncertainty on the object identity and pose at the early inference stage, the robot often comes into contact with the object in unexpected timing and locations. 

\begin{figure}[thpb]
  \centering
  \includegraphics[width=\linewidth]{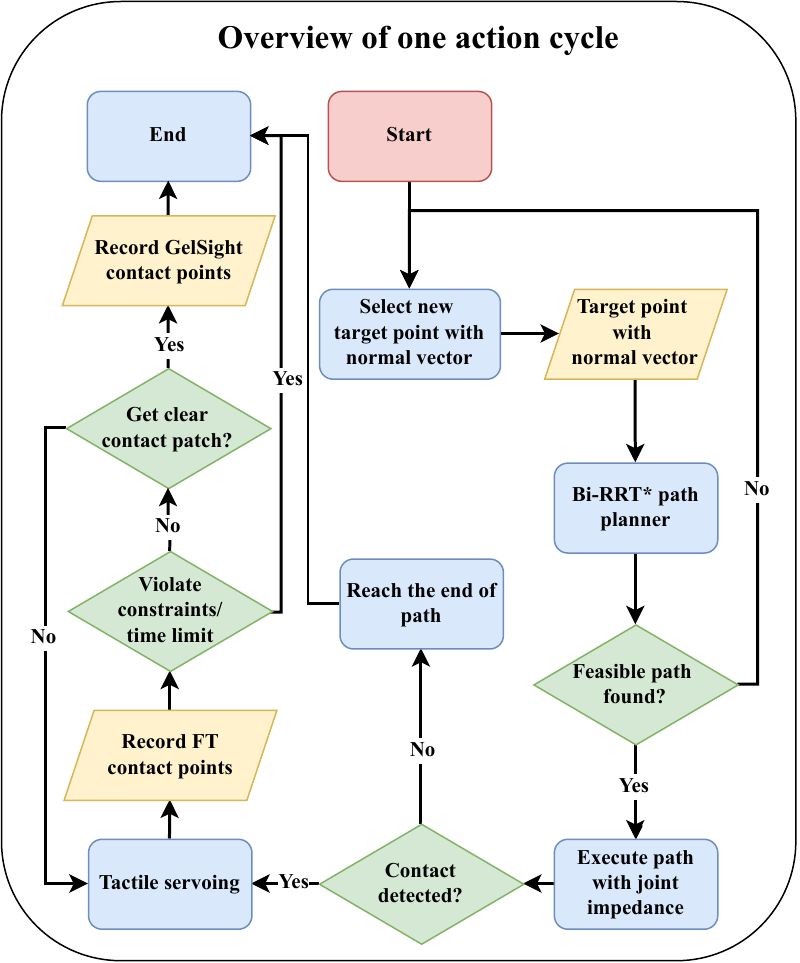}
  \caption{An overview of one action cycle in the planning and control stack. The target selection scheme proposes target points with normal vectors as inputs to the Bi-RRT* motion planner until a feasible path to the target is found. The path is executed with joint impedance control. If a contact is detected by the F/T sensor during execution, it triggers tactile servoing to align the GelSight sensor with the contact surface.}
  \label{fig:controller_logic}
\end{figure}

To collect data effectively, we design a motion planning and control stack, whose logic is shown in (Fig.~\ref{fig:controller_logic}). For each action cycle, the stack finds a feasible target point to explore, plans and executes the trajectory while handling contacts and data collection with a wrist force/torque sensor and a GelSight Mini sensor mounted on the robot arm. More detailed descriptions are given in the following subsections.

\subsubsection{Path Planner}
For trajectory planning, we implement a Bi-directional RRT* (Bi-RRT*) planner in PyBullet for the Franka Emika Panda robot arm. The planning scene includes an approximate lab environment and the mesh of the current MAP object hypothesis (class and pose), updated at each time step. Given the target point $\overline{\mathbf{x}}_{t+1}$ and its surface normal $\overline{\mathbf{n}}_{t+1}$, the planner generates a collision-free joint-space path to a pre-touch configuration.

\subsubsection{Trajectory planning}\label{subsubsec:traj_plan}
\begin{figure}[thpb]
  \centering
  \includegraphics[width=\linewidth]{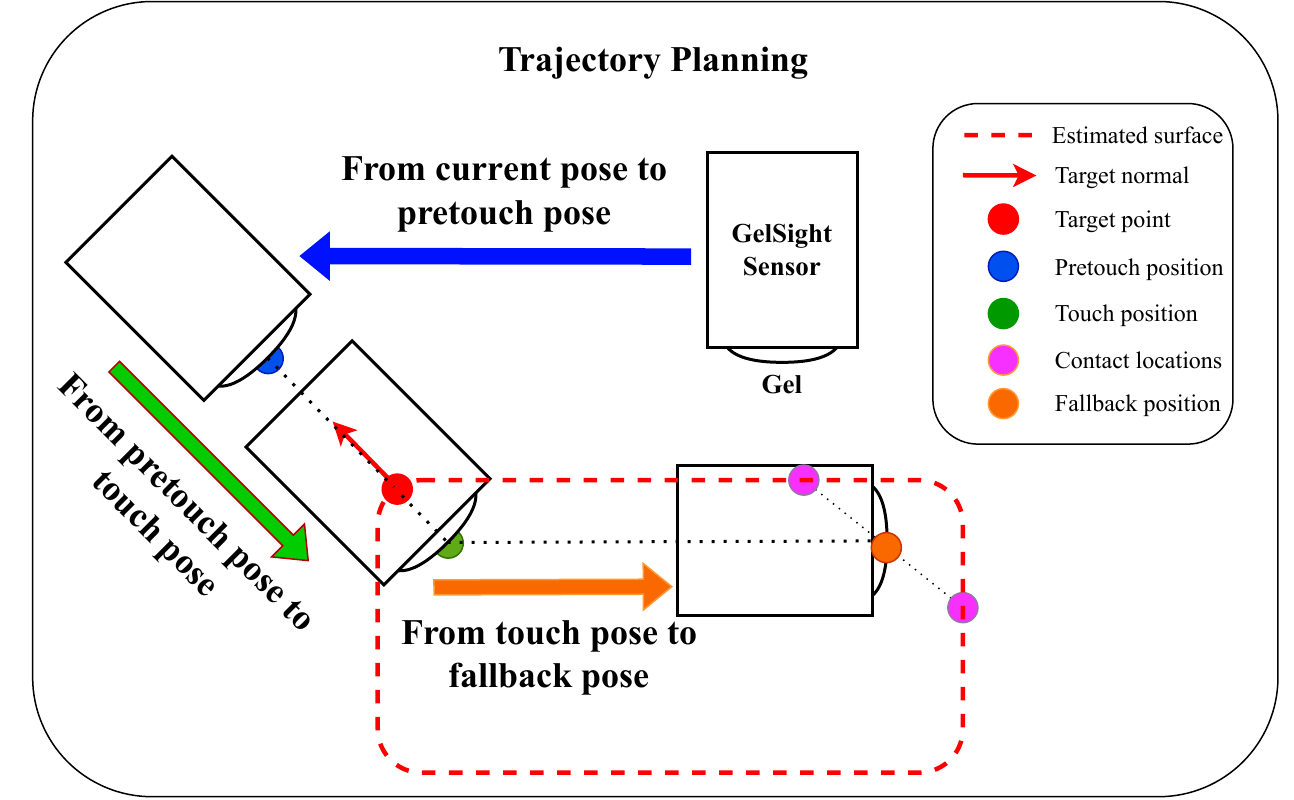}
  \caption{ An overview of the trajectory planning process given a target point with a normal direction. The trajectory consists of three segments: from the current pose to a pre-touch pose where the sensor aligns with the normal vector and faces directly toward the target point; from the pre-touch pose moving long the opposite normal direction to a touch pose to establish contact with the estimated surface; from the touch pose to the fallback pose located at the centroid of the existing contact points.}
  \label{fig:trajectory_planning}
\end{figure}
Given a target $(\mathbf{x},\mathbf{n})$, we compute a pre-touch pose, a touch pose, and a fallback pose, and solve the inverse kinematics (damped least squares) to obtain the corresponding joint configurations. The pre-touch pose places the sensor $3\,\mathrm{cm}$ from $\mathbf{x}$ along $-\mathbf{n}$ and orients the gel normal $\mathbf{n}_G$ towards $-\mathbf{n}$. Since position and pointing direction do not uniquely determine a 6D pose, we discretize a rotation angle $\theta_n$ about $\mathbf{n}_G$ to generate multiple candidates. The pre-touch pose $(\mathbf{t}_p,\mathbf{u}_p)$ is

\begin{align}
\mathbf{t}_p &= \mathbf{x}-0.03\,\mathbf{n},\\
\theta_p &= \mathrm{atan2}\!\left(\|\mathbf{n}_G\times(-\mathbf{n})\|,\ \mathbf{n}_G\cdot(-\mathbf{n})\right),\\
\mathbf{u}_r &= (\cos(\frac{\theta_p}{2}),\sin(\frac{\theta_p}{2})\cdot\mathbf{r}_p)\\
    \mathbf{u}_n &= (\cos(\frac{\theta_n}{2}),\sin(\frac{\theta_n}{2})\cdot\mathbf{n}_{G})\\
\mathbf{r}_p &= \frac{\mathbf{n}_G\times(-\mathbf{n})}{\|\mathbf{n}_G\times(-\mathbf{n})\|},\qquad
\mathbf{u}_p = \mathbf{u}_r\,\mathbf{u}_n\,\mathbf{u},
\end{align}
When $\|\mathbf{n}_G\times(-\mathbf{n})\|=0$, we choose any unit axis $\mathbf{r}_p\perp \mathbf{n}_G$ (e.g., by projecting a random vector onto the orthogonal complement of $\mathbf{n}_G$ and normalizing).

The touch pose advances $5\,\mathrm{cm}$ from the pre-touch pose along $-\mathbf{n}$ (contact expected). The fallback pose uses the centroid of previous contacts as the target position and the direction from the expected touch position toward the target position as the approach direction. The resulting trajectory has three segments. (i) A collision-free segment planned by Bi-RRT* from the current joint configuration to the pre-touch configuration. (ii) A short local approach from pre-touch to touch, where contact is expected if the estimated class and pose are correct; this segment is generated by Cartesian linear interpolation with inverse kinematics. (iii) If no contact is detected at the touch pose and a feasible fallback configuration is found, the fallback segment is executed by joint-space interpolation from touch to the fallback configuration, steering the sensor toward a region where contact is more likely (for convex objects, the fallback position lies inside the object). In this manner, even without contact, the collision-free motion provides free-space information to update the belief over object class and pose.

\subsubsection{Trajectory execution}
As collisions could take place at any phase of the trajectory execution, compliance in the controller is necessary to prevent damaging the robot hardware. Thus, trajectories are executed with a joint impedance controller for compliance.
\begin{align}
     \tau &= \mathbf{K_J} (\mathbf{q_d}-\mathbf{q)} - \mathbf{D_J} \dot{\mathbf{q}} + \mathbf{C}(\mathbf{q},\dot{\mathbf{q}}) +\mathbf{G}(\mathbf{q})\\
     \mathbf{K_J} &= \text{diag}\left([k_1,k_2,k_3,k_4,k_5,k_6,k_7]\right)\\
     \mathbf{D_J } &= \text{diag}\left([d_1,d_2,d_3,d_4,d_5,d_6,d_7]\right)
\end{align}
where $\mathbf{\tau}$ is the commanded torque, $\mathbf{q}$ and $\mathbf{q_d}$ are the current joint configuration and the desired joint configuration, respectively. The joint stiffness matrix and the joint damping matrix are denoted as $\mathbf{K_J }$ and $\mathbf{D_J}$ respectively, $\mathbf{C}(\mathbf{q},\dot{\mathbf{q}})$ is the Coriolis compensation term, $\mathbf{G}(\mathbf{q})$ is the gravity compensation term.

The controller tracks the trajectory at $0.05$ seconds per waypoint. If the joint-space tracking error $|\mathbf{q}-\mathbf{q}_d|$ exceeds a predefined threshold, the current waypoint is held until the error falls below the threshold. If the error remains above the threshold for more than $3$ seconds, the system treats it as a body/environment collision, terminates the execution of the current trajectory, and triggers the recovery mode (see corner-case handling). This error bound also limits the commanded torques under joint impedance control, preventing emergency stops. If contact is detected during trajectory execution, tactile servoing is activated.

\subsubsection{Tactile Servoing}
\begin{figure}[thpb]
  \centering
  \includegraphics[width=\linewidth]{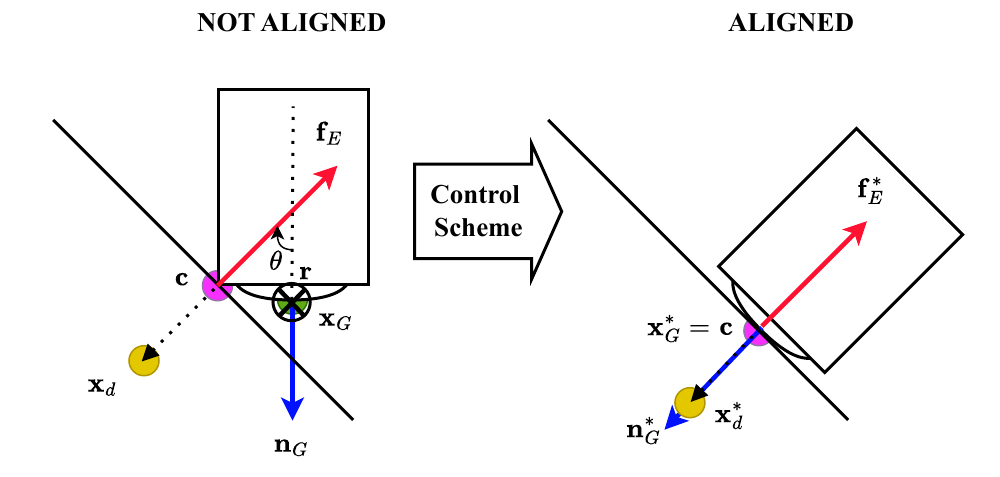}
  \caption{Control scheme for tactile servoing: at each step of the control loop (60Hz), we compute a target gel-center position $\mathbf{x}_d$ and a rotation $(\theta,\mathbf{r})$ to align the gel normal $\mathbf{n}_G$ (blue) with the opposite external force direction $-\mathbf{f}_E$ (red) while maintaining contact. The commands are derived from the estimated contact location $\mathbf{c}$, current gel center $\mathbf{x}_G$ (green), $\mathbf{n}_G$, and $\mathbf{f}_E$. A Cartesian impedance controller outputs joint torques $\tau$ to drive $\mathbf{x}_G\rightarrow\mathbf{x}_d$ and $\mathbf{n}_G\rightarrow-,\mathbf{f}_E$, ideally converging to a force equilibrium (superscript $^{*}$).}
  \label{fig:alignment_controller}
\end{figure}

Since the estimated shape may not perfectly match the true object, the vision-based tactile sensor can collide in unintended configurations, e.g., with incomplete gel contact or a contact patch located near the gel boundary, resulting in no patch or a small patch. To address this, we propose a tactile servoing scheme that aligns the gel with the object surface and centers the contact patch.

We compute the desired orientation $\mathbf{q}_d$ by rotating the gel normal vector $\mathbf{n}_G$ toward the reaction force direction, approximated by the negative external force $-\mathbf{f}_E$. Specifically, we find a rotation axis $\mathbf{r}$ and rotation angle $\theta$ such that a single rotation aligns $\mathbf{n}_G$ with $-\mathbf{f}_E$. The axis and angle are obtained from the cross and dot products of $\mathbf{n}_G$ and $-\mathbf{f}_E$:
\begin{align}
\theta &= \mathrm{atan2}\!\left(\left\|\mathbf{n}_{G} \times (-\mathbf{f}_{E})\right\|,\ \mathbf{n}_{G} \cdot (-\mathbf{f}_{E})\right),\\
\mathbf{r} &= \frac{\mathbf{n}_{G} \times (-\mathbf{f}_{E})}{\left\|\mathbf{n}_{G} \times (-\mathbf{f}_{E})\right\|},
\label{eqn:rotation_axis_2}
\end{align}
When $\|\mathbf{n}_{G} \times (-\mathbf{f}_{E})\|=0$, the rotation axis in \eqref{eqn:rotation_axis_2} is ill-defined. In this case, we choose any unit vector $\mathbf{r}\perp \mathbf{n}_{G}$, constructed using the same procedure as in Sec.~\ref{subsubsec:traj_plan}. To avoid rapid motions, we clip the rotation angle to $45^\circ$:
\begin{equation}
\theta = \min\!\left(\mathrm{atan2}\!\left(\|\mathbf{n}_{G}\times(-\mathbf{f}_{E})\|,\ \mathbf{n}_{G}\cdot(-\mathbf{f}_{E})\right),\ \frac{\pi}{4}\right).
\end{equation}
To maintain contact during alignment, we set the target gel-center position $\mathbf{x}_d$ a distance $\alpha$ along the reaction-force direction $-\mathbf{f}_E$ from the estimated contact location $\mathbf{c}$:

\begin{equation}
\mathbf{x}_{d} = \mathbf{c}-\alpha\frac{\mathbf{f}_{E}}{\|\mathbf{f}_{E}\|}.
\end{equation}
The offset $\alpha$ regulates the interaction force: overly large forces may damage the gel, whereas overly small forces may not produce sufficient indentation. We therefore adapt $\alpha$ based on the measured force magnitude. Given a desired static force magnitude $\|\mathbf{f}^{*}_{E}\|$ and translational stiffness $k_t$, we set

\begin{equation}
    \alpha = \frac{k_f}{k_{t}}(||\mathbf{f}^{*}_{E}||-||\mathbf{f}_{E}||)
\end{equation}
where $k_f$ is a force-servoing gain. We choose $k_f>k_t$ to compensate for the fact that the measured wrench is influenced by both translation and rotation during alignment; if $k_f$ is small, it will not be able to reduce the interaction force effectively.
Additionally, once a contact patch is obtained by the GelSight sensor, the controller replaces the estimated contact location of the force torque sensor with the contact patch center on the tactile image to center the contact patch towards on the tactile image.
The Cartesian-impedance tactile servoing law is
\begin{align}
     \mathbf{\epsilon} &= \left[\mathbf{x}_{d}-\mathbf{x}_{G}, \theta\mathbf{r}\right]\\
     \tau &= \mathbf{J}_{G}^{T}(\mathbf{K}\mathbf{\epsilon} - \mathbf{D}\mathbf{J}_{G}\dot{\mathbf{q}}) + \mathbf{C}(\mathbf{q},\dot{\mathbf{q}}) +\mathbf{G}(\mathbf{q})\\
     \mathbf{K} &= \text{diag}\left([k_t,k_t,k_t,k_r,k_r,k_r]\right)\\
     \mathbf{D} &= 2\sqrt{\mathbf{K}} 
\end{align}
where $\mathbf{\tau}$ is the commanded torque and $\mathbf{q}$ is the current joint configuration;  $\mathbf{J}_{G}$ is the Jacobian matrix calculated at the gel center $\mathbf{x}_{G}$. $\mathbf{K}$ and $\mathbf{D}$ are the Cartesian stiffness matrix and the Cartesian damping matrix respectively, while $\mathbf{C}(\mathbf{q},\dot{\mathbf{q}})$ and $\mathbf{G}(\mathbf{q})$ denote Coriolis and gravity compensation terms, respectively. The damping in each direction is chosen heuristically to yield critical damping for the corresponding virtual mass--spring--damper system, assuming a unit virtual mass of $1\,\mathrm{kg}$.

 If a contact patch is observed from the tactile sensor for more than 0.5 seconds, the contact patch will be processed and passed to the Bayesian framework as input. The controller then retracts the GelSight sensor away from the object surface along its pointing direction. 

\subsubsection{Corner cases handling}
To improve robustness during autonomous exploration, we implement simple recovery behaviors for common planning/execution failures: (i) proximity to joint limits or self-collision during execution or tactile servoing, (ii) unintended body/environment collisions due to model mismatch, (iii) failure to obtain a stable tactile patch within a time limit, (iv) repeated sampling of the same region (e.g., when the start pose is too close to the object), and (v) infeasible planning queries caused by slight penetration between the robot and the estimated object mesh in the planner.

In cases (i)--(ii), the robot aborts and retraces the executed trajectory in reverse to return to a safe start configuration. In case (iii), tactile servoing is terminated, the sensor retracts, and the collected free-space and F/T contact cues are still integrated into the Bayesian update. In case (iv), the robot moves to a predefined reset configuration (collision-free and $>50\,\mathrm{cm}$ above the table) before resuming exploration. In case (v), the planner resolves small penetrations by shifting the estimated object pose away from the robot in increments of $1\,\mathrm{mm}$ until PyBullet reports no penetration.

 \begin{table*}[t]
\centering
\small 
\setlength{\tabcolsep}{4pt} 
\begin{tabular}{lcccccccc}
\toprule
\textbf{\#Actions} &
\multicolumn{4}{c}{\textbf{Simulation}} &
\multicolumn{4}{c}{\textbf{Real world}} \\
\cmidrule(lr){2-5}\cmidrule(lr){6-9}
& \textbf{FT} & \textbf{FT+GS} & \textbf{FT+FS} & \textbf{FT+GS+FS}
& \textbf{FT} & \textbf{FT+GS} & \textbf{FT+FS} & \textbf{FT+GS+FS} \\
\midrule
\textbf{Classification} &  $4.3\pm3.2$ & $4.1\pm2.7$ & $3.6\pm2.0$ & $\bm{3.4\pm1.9}$ & $6.6\pm4.7$ & $6.7\pm4.6$ & $5.0\pm3.1$ & $\bm{4.2\pm2.2}$ \\
\textbf{Pose} & $6.6\pm3.7$  & $6.1\pm3.5$  & $5.6\pm2.4$  & $\bm{4.7\pm2.1}$  & $8.6\pm4.5$  & $7.9\pm4.4$  & $6.3\pm2.8$  & $\bm{5.3\pm1.7}$  \\
\bottomrule
\end{tabular}
\caption{Mean and standard deviation of action cycles needed to obtain the correct class and the desired pose error on average.}
\label{tab:sim_real_methods}
\end{table*}
\begin{figure}[t]
  \centering
  \includegraphics[width=\linewidth]{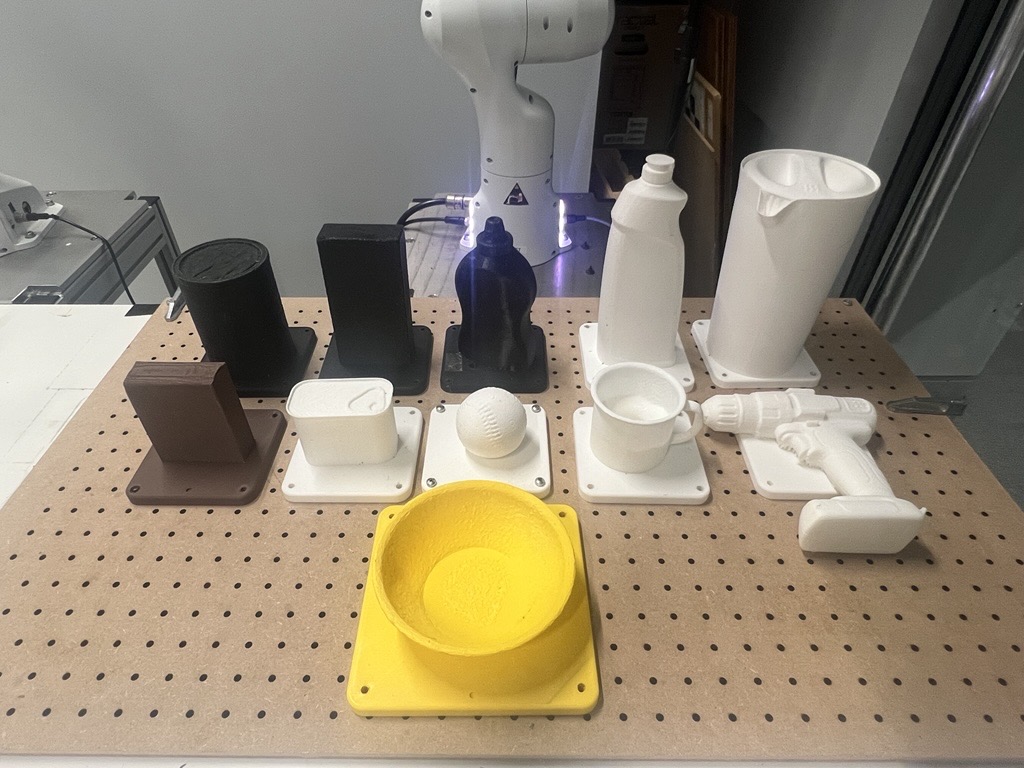}\hfill
  \caption{The 3D-printed YCB objects used in the experiments.}
  \label{fig:object_set}
\end{figure}
\begin{figure*}
  \centering
  \includegraphics[width=\textwidth]{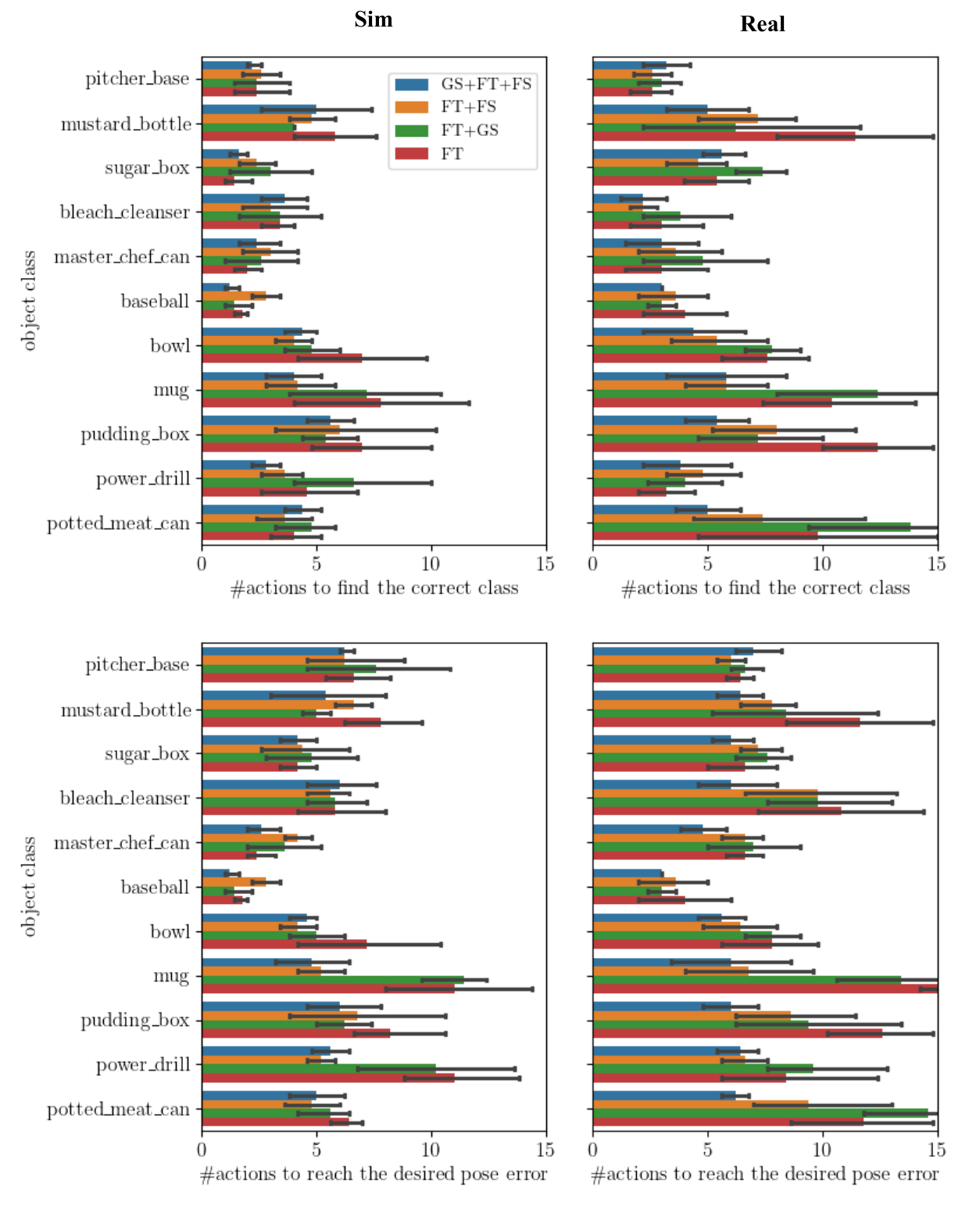}\hfill
  \caption{Class-wise number of actions required to find the correct class and reach the desired pose estimation error in simulation and real-world experiments.}
  \label{fig:class_wise_histo}
\end{figure*}
\begin{figure*}[t]
  \centering
  \includegraphics[width=\textwidth]{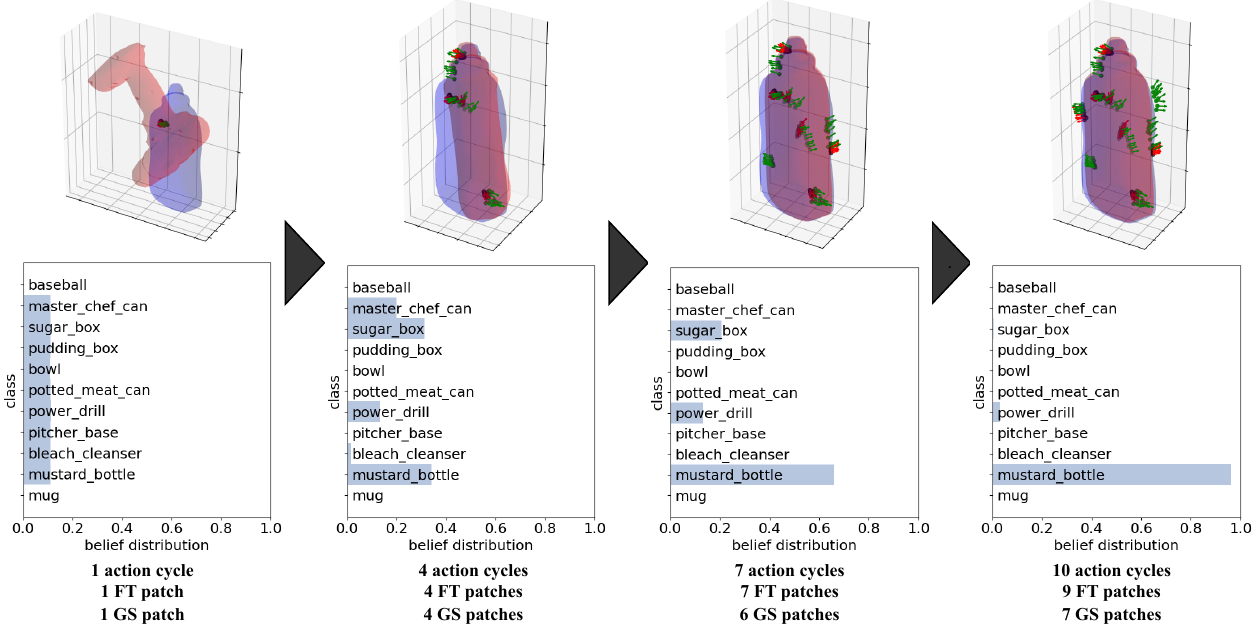}
  \caption{An example of how the estimated class and pose evolves as more data are collected, taken from one of the real-world experiments on the mustard bottle. The top row shows the ground truth object class and pose in blue and the MAP estimation in red, with the FT contact points (green) and GelSight contact points (black + red). The bottle row shows the belief distribution on the object class. From left to right, as more data are collected, the framework gradually became more confident about the object class and refined the object pose.}
  \label{fig:inference_evolution}
\end{figure*}

\begin{table}[t]
\centering
\small 
\setlength{\tabcolsep}{4pt} 
\begin{tabular}{lcccccc}
\toprule
\textbf{Distance (mm)} &
\multicolumn{3}{c}{\textbf{Simulation}} &
\multicolumn{3}{c}{\textbf{Real world}} \\
\cmidrule(lr){2-4}\cmidrule(lr){5-7}
& \textbf{mean} & \textbf{stdv} & \textbf{max}
& \textbf{mean} & \textbf{stdv} & \textbf{max} \\
\midrule
\textbf{GelSight} & 1.2 & 0.8 & 6.5 & 3.3 & 2.0 & 17.7  \\
\midrule
\textbf{Force torque} & 0.7 & 0.8 & 6.4 & 4.5 & 2.4 & 36.1 \\
\bottomrule
\end{tabular}
\caption{Mean, standard deviation and maximum of the distance from the recorded contact points to the ground truth surface in simulation and in real-world.}
\label{tab:sim_real_noise}
\end{table}

\begin{figure}[t]
  \centering
  \includegraphics[width=\linewidth]{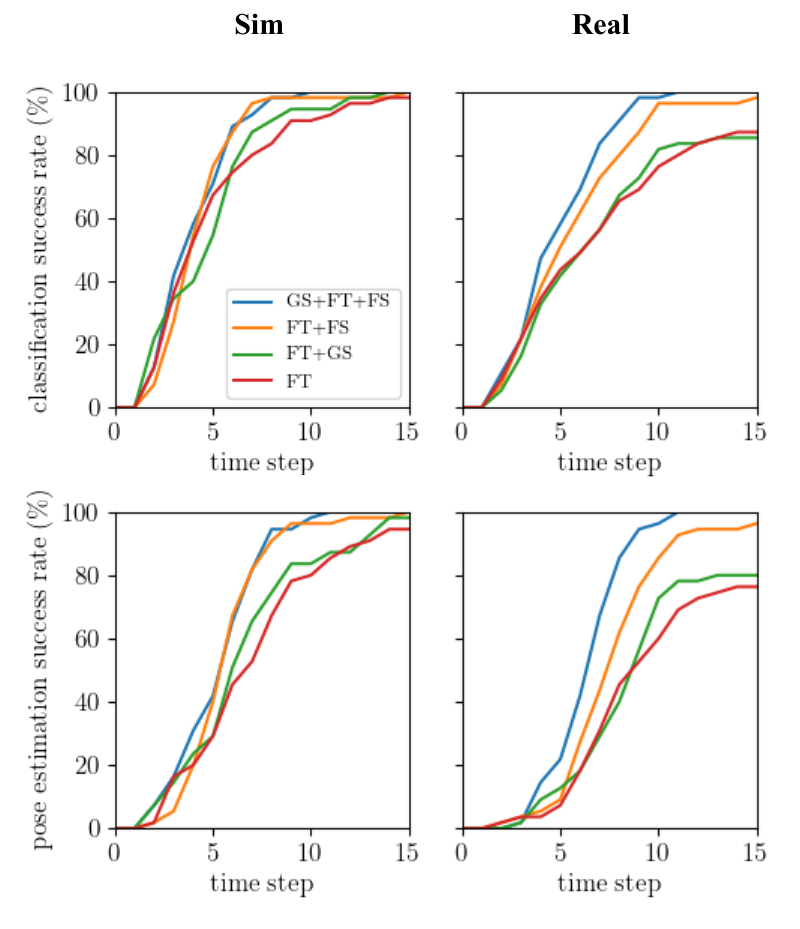}
  \caption{Classification success rate and pose estimation success rate over time step in the simulated and real-world experiments.}
  \label{fig:accuracy_over_time}
\end{figure}
\section{Experimental Results}

We evaluate our approach in both simulation and the real world. Performance is measured by (i) the number of actions required to correctly recognize the object (i.e., after which the estimated class remains consistent with the ground truth), and (ii) the number of actions required to achieve a consistently accurate pose estimate. Pose accuracy is quantified using the ADD-S metric~\cite{xiang2018posecnnconvolutionalneuralnetwork}, which computes the mean Euclidean distance between closest 3D model points under the predicted and ground-truth poses; we report success if ADD-S drops below $6\,\mathrm{mm}$ and remains below $6\,\mathrm{mm}$ for the rest of the trial.

\textbf{Real-world setup.} Experiments are conducted in our lab using a Franka Emika Panda robot arm, a Bota Systems SensONE F/T sensor, and a GelSight Mini sensor. The GelSight sensor is mounted to the F/T sensor via a custom flange, and the F/T sensor is mounted on the robot wrist, see Fig.~\ref{fig:scheme}.

\textbf{Simulation setup.} We build a digital twin of the real setup in PyBullet~\cite{coumans2016pybullet}, including the table, lab walls, and the Panda arm with the F/T and GelSight sensors. Objects are initialized with the same poses as in the real experiments. F/T contact points are simulated in PyBullet, while GelSight contact patches are generated using TACTO~\cite{wangTACTOFastFlexible2022}.

\textbf{Objects and protocol.} We evaluate our framework on 11 objects from the YCB model set~\cite{7254318}, selected to span diverse shapes while including similarly sized objects with comparable local surface structure (e.g., two boxes and two cans), which are shown in Fig.~\ref{fig:object_set}. Following~\cite{caddeo2024sim2real}, we 3D-print the objects to prevent undesired motion during contact. For each object, we run 5 trials with the same object pose but randomized initial contact locations, resulting in different exploration sequences. Each trial lasts for 15 action cycles (time steps), where one cycle corresponds to executing a full trajectory (trajectory execution plus tactile servoing).

\textbf{Baselines.} To assess the contribution of each sensing source, we compare our full method with three ablations: using F/T sensor only, using F/T sensor and GelSight sensor, using F/T sensor and free-space map, abbreviated as \textbf{FT} only, \textbf{FT+GS}, and \textbf{FT+FS} respectively.

\textbf{Dataset:}
During real-world experiments, we record at each time step the ground-truth object class and pose, the MAP class and pose, F/T-based contact points, GelSight contact points, and the free-space map. The resulting dataset captures the noise characteristics of our hardware setup and enables offline evaluation and comparison of inference methods under identical sensing conditions.

For the \textbf{FT} only, \textbf{FT+GS}, and \textbf{FT+FS} ablations in the real-world setting, we use the recorded dataset and replay only the corresponding modalities as input to the Bayesian inference module. In simulation, these ablations are executed from scratch rather than via replay. The key difference is that replay fixes the contact and free-space observations at each time step to the recorded sequence, whereas running from scratch allows the active exploration policy to guide the data collection.

\textbf{Sim2Real noise} For completeness, we report the noise levels of contact points in both simulation and real-world experiments, as shown in Table~\ref{tab:sim_real_noise}. We measure noise as the distance from each recorded contact point to the nearest vertex on the ground-truth object surface. In the real-world setting, this metric also includes calibration-induced errors. Consequently, if the calibrated ground-truth pose is imperfect, the true sensor noise may differ from the reported value. The same caveat applies to the reported pose error, which is computed as ADD between the MAP estimation and the calibrated ground truth. Since calibration error affects both the noise metric and the pose-error metric, our results should be interpreted as the achieved object recognition and pose estimation performance under the corresponding reported noise level.

\begin{table*}[t]
\centering
\small 
\setlength{\tabcolsep}{4pt} 
\begin{tabular}{lcccccccc}
\toprule
\textbf{ADD-S (mm)} &
\multicolumn{4}{c}{\textbf{Simulation}} &
\multicolumn{4}{c}{\textbf{Real world}} \\
\cmidrule(lr){2-5}\cmidrule(lr){6-9}
& \textbf{FT} & \textbf{FT+GS} & \textbf{FT+FS} & \textbf{FT+GS+FS}
& \textbf{FT} & \textbf{FT+GS} & \textbf{FT+FS} & \textbf{FT+GS+FS} \\
\midrule
\textbf{mean}$\pm$\textbf{stdv}& $1.8\pm1.3$  & $1.8\pm0.9$  & $\bm{1.6\pm0.6}$  & $1.6\pm0.7$  & $6.1\pm8.6$  & $7.8\pm11.8$  & $3.3\pm1.1$  & $\bm{2.9\pm0.8}$  \\
\textbf{max} & $9.3$  & $4.5$  & $\bm{2.9}$  & $3.4$  & $48.8$  & $49.8$  & $9.1$  & $\bm{4.9}$  \\
\bottomrule
\end{tabular}
\caption{Mean and standard deviation of the final pose estimation error (after 15 action cycles) measured by ADD for simulation and real-world experiments average over all object classes.}
\label{tab:sim_real_end_pose}
\end{table*}

\subsection{Results}  

Table~\ref{tab:sim_real_methods} reports the mean number of action cycles required to reach the correct class and to achieve the target pose accuracy, averaged over objects and trials. An F/T contact is recorded only when contact occurs, and a GelSight patch is recorded only when alignment succeeds. Overall, fusing complementary cues reduces the number of action cycles and improves estimation accuracy. Adding GelSight observations to F/T yields only modest gains, whereas incorporating free-space (FS) constraints substantially improves performance by pruning the hypothesis space. Relative to \textbf{FT} only, \textbf{FT+FS} improves the metrics by approximately 15--20\% in both simulation and real experiments, and fusing all modalities (\textbf{FT+GS+FS}) yields consistent gains above 20\% (up to $\sim$30\%). The reduced standard deviation further indicates that multimodal fusion improves stability and reliability.

Fig.~\ref{fig:class_wise_histo} provides a class-wise breakdown of the number of action cycles for correct classification (top) and pose accuracy (bottom). Integrating all three modalities is consistently the best or second-best method across objects in both simulation and the real world, with similar relative behavior across scenarios. Objects with similar shapes/sizes to others in the set (e.g., \textit{mustard\_bottle}, \textit{mug}, \textit{sugar\_box}, \textit{pudding\_box}, \textit{bowl}) typically require more actions to disambiguate, while distinctive objects (e.g., \textit{baseball}) require fewer. The \textit{potted\_meat\_can} is an outlier; we analyze this case later in this section.

To demonstrate how active inference evolves over time, an example is given in Fig.~\ref{fig:inference_evolution}, as more information was made available through active exploration, the uncertainty in the belief distribution dropped and the estimated pose was refined.
Fig.~\ref{fig:accuracy_over_time} shows classification and pose-estimation success rates as a function of action cycles. In real-world experiments, \textbf{FT+GS+FS} reaches 100\% success rate in both metrics after 11 cycles, outperforming all ablations. The strong gap between methods with FS (\textbf{FT+FS}, \textbf{FT+GS+FS}) and those without (\textbf{FT}, \textbf{FT+GS}) indicates that FS is crucial for resolving class and pose ambiguity. Similar trends hold in simulation, with smaller gaps due to differences in experimental protocol (replay vs.\ online exploration) and the lower noise level of simulated F/T contacts.

Table~\ref{tab:sim_real_end_pose} reports the final pose error after 15 cycles. In simulation, \textbf{FT+FS} achieves the lowest final error, while in real experiments \textbf{FT+GS+FS} performs best and \textbf{FT+FS} is second due to one misclassification case. Notably, in the real-world replay ablations, \textbf{FT+GS} can yield a larger final pose error than \textbf{FT} on some objects (e.g., \textit{potted\_meat\_can} and \textit{mug}). Since replay fixes the contact sequence, removing modalities can introduce ambiguities that the system cannot actively resolve by selecting new informative targets; in such cases, adding GelSight may bias the posterior toward hypotheses that explain sparse local contacts well but correspond to an incorrect class or a misaligned pose. To verify that this effect is not caused by poor contact fitting, we compute the mean distance from observed contacts to the final MAP surface; in the worst case, the mean distances for \textbf{FT+GS+FS}, \textbf{FT+GS}, \textbf{FT+FS}, and \textbf{FT} are $5.5\,\mathrm{mm}$, $5.0\,\mathrm{mm}$, $5.0\,\mathrm{mm}$, and $4.8\,\mathrm{mm}$, respectively, indicating comparable fits to the contact data. Therefore, the larger pose error in \textbf{FT+GS} is mainly attributable to residual class/pose ambiguity under sparse contact-only evidence.
The performance gap between simulated experiments and the real-world experiments in the \textbf{FT+GS+FT} condition is due to the non-comparable level of noise between the two scenarios described in Table~\ref{tab:sim_real_noise}. 
 The larger noise level in real-world experiments is likely the main cause for the performance drop compared to the low-noise scenario in the simulation as the initial conditions were kept as close as possible between the simulated and the real-world experiments.

Finally, as an index of the planning and control stack performance in real experiments, $91.5\%$ of action cycles produced an F/T contact (755/825), and $72.8\%$ of contacts resulted in successful GelSight alignment (550/755). The failure of alignment is caused by either the robot getting close to joint limit and self-collision configurations, or taking too long to stabilize the GelSight sensor around the contact surface.



\section{Discussion}
\subsection{Role of each modality}
As shown in the Results section, fusing complementary information sources achieved more robust object recognition and pose estimation. Tactile observations are inherently local and sparse; resolving class--pose ambiguity therefore requires accumulating informative evidence over multiple interactions.
In practice, relying on GelSight alone is not always sufficient. Accurate GelSight measurements require the sensor to be well aligned with the contact surface, and this can be difficult to guarantee under kinematic constraints (e.g., joint limits and self-collision) and model mismatch. In contrast, wrist force/torque (F/T) sensing and motion-derived free-space information remain available even when GelSight alignment is unsuccessful. In our framework, F/T contact estimates provide coarse but broadly available constraints that are useful for proposing and scoring hypotheses, while the free-space map acts as a global constraint that rules out hypotheses inconsistent with collision-free motion. This explains the strong gains observed when incorporating free-space information.
When alignment succeeds, GelSight provides more precise local geometric cues (contact patch and normal estimates) than F/T alone. These higher-fidelity measurements improve sample efficiency by enabling better hypothesis proposals and more accurate pose refinement, and they are particularly helpful for reaching the target pose threshold with fewer action cycles. 

\subsection{The role of the planning and control stack}
The planning and control stack is a key enabler for fully autonomous active exploration under uncertainty. It allows the robot to execute reachability-aware target motions, remain compliant during contact, and handle both expected and unexpected collisions without interrupting the exploration process. Importantly, the stack ensures that informative measurements are collected reliably: it provides contact events for F/T-based estimation, enables stable GelSight alignment when possible through tactile servoing, and continuously accumulates free-space evidence from collision-free motion. These capabilities directly contribute to the overall performance observed in both simulation and real-world experiments.

\subsection{Limitations and Future work}

This work assumes that (i) the object is rigid and remains stationary during interaction and (ii) only a single object is present in the scene. Relaxing these assumptions---e.g., to handle object motion, articulated objects, or multi-object scenes---is an important direction for future work.
The computational cost of the current inference procedure scales approximately linearly with the number of object priors. Improving scalability (e.g., via hierarchical retrieval, coarse-to-fine hypothesis selection, or learned proposal distributions) is another promising direction.
Finally, while we focus on contact-based perception in this study, the Bayesian formulation naturally supports additional sensing sources. Integrating vision cues such as point clouds, pose tracking, and segmentation could further improve robustness in cluttered scenes and enable more general manipulation tasks. Beyond scaling the approach by collecting larger datasets, an important direction is to extend the framework to recognize and reconstruct novel objects beyond the prior set.

\section{Conclusion}
In this paper, we addressed the problem of active object recognition and pose estimation by fusing information from a GelSight sensor, a wrist force/torque sensor and robot motion under a unified Bayesian framework, with a planning and control stack for efficient active data acquisition. 
Through extensive experiments in both simulation and real-world lab environments, we show that our system can recognize the object within 4 action cycles and estimate its pose within 6 action cycles on average.
We believe that our system is a step toward general-purpose perception for robots that can exploit the full spectrum of contact-derived cues and remain effective when other sensing signals are unreliable or unavailable.
\bibliographystyle{IEEEtran}
\bibliography{citing_papers}

 \begin{IEEEbiography}[{\includegraphics[width=1in,height=1.25in,clip,keepaspectratio]{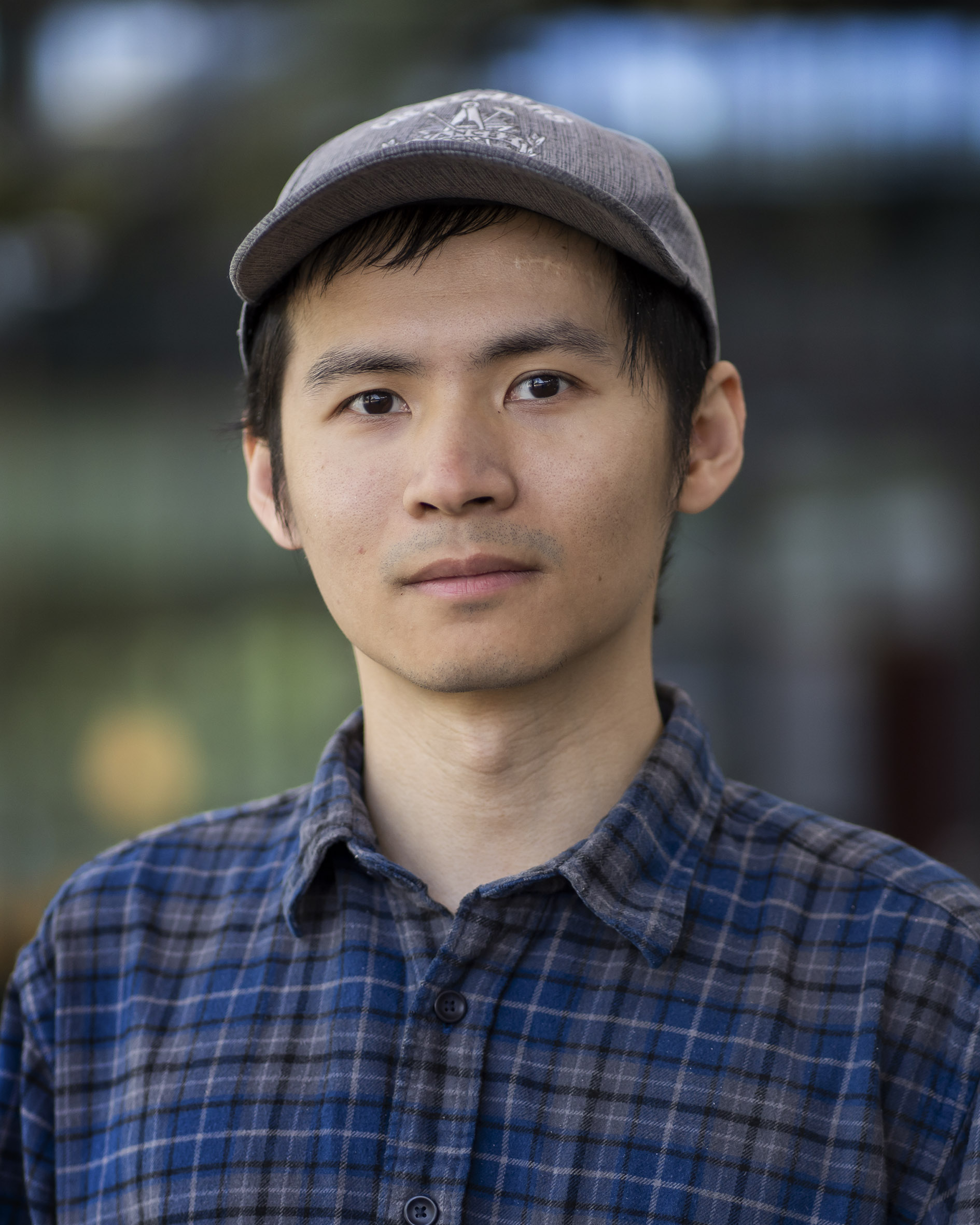}}]{Haodong Zheng} received his B.Eng. Degree in Mechatronic Engineering in 2017 at the Shantou University, China, and his M.Sc. Degree in Complex Adaptive Systems in 2022 at the Chalmers University of Technology, Sweden. He is currently a PhD candidate in the Human Technology Interaction Group at the Eindhoven University of Technology, The Netherlands. His research interests include robotic visuo-tactile perception and manipulation through Deep Learning and Bayesian methods. 
\end{IEEEbiography}
\vskip 0pt plus -1fil

\begin{IEEEbiography}
[{\includegraphics[width=1in,height=1.25in,clip,keepaspectratio]{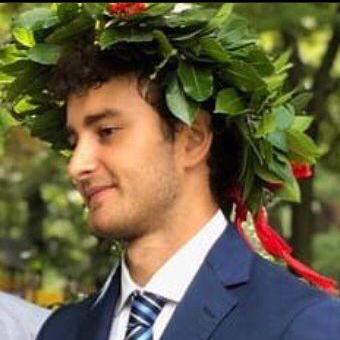}}]{Gabriele Mario Caddeo} received his B.S. degree in biomedical engineering, in 2018, the M.S. degree in electronic engineering, in 2020, from the Polytechnic University of Turin, and the PhD in Robotics in 2025, from the University of Genoa and the Italian Institute of Technology, in the Humanoid Sensing and Perception Lab, under the supervision of Lorenzo Natale. He was previously a Visiting Researcher at the Technical University of Munich and the University of Maryland. He is currently a postdoctoral researcher at the same lab.  His research interests include generative AI and sensorimotor policies for perception and manipulation with sensor fusion through deep learning and classical methods
\end{IEEEbiography}
\vskip 0pt plus -1fil

\begin{IEEEbiography}[{\includegraphics[width=1in,height=1.25in,clip,keepaspectratio]{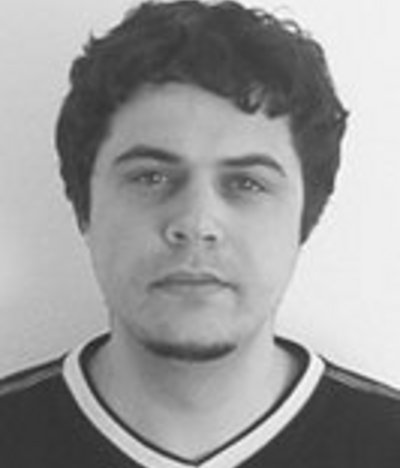}}]
{Andrei C. Jalba} received his Ph.D. degree in 2004 from the Institute for
Mathematics and Computing Science of the University of Groningen, The
Netherlands. Currently, he is an assistant professor in visualization and computer graphics at the Eindhoven University of Technology, the Netherlands. 
His expertise includes multi-scale shape representation, processing 
and reconstruction, physics-based material simulation and visual,
scientific and multi-core computing.
\end{IEEEbiography}

\vskip 0pt plus -1fil
\begin{IEEEbiography}[{\includegraphics[width=1in,height=1.25in,clip,keepaspectratio]{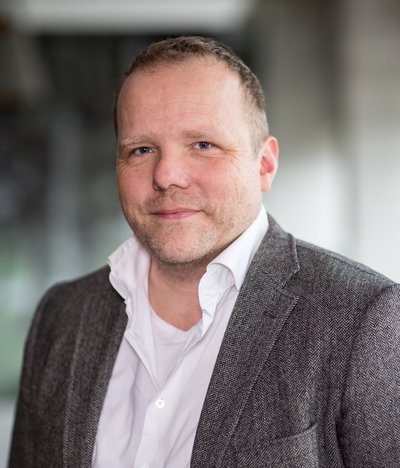}}]
{Wijnand IJsselsteijn} is a full professor of Cognition and Affect in Human-Technology Interaction at Eindhoven University of Technology (TU/e). He has an active research program on the impact of media technology on human psychology, and the use of psychology to improve technology design. His focus is on conceptualizing and measuring human experiences in relation to digital environments (immersive media, serious games, affective computing, personal informatics) in the service of human learning, health, and wellbeing. He has a keen interest in the relation between data science, AI and psychology, and works on technological innovations (such as sensor-enabled mobile technologies, virtual environments) that make possible novel forms of human behavior tracking, combining methodological rigor with ecological validity.
\end{IEEEbiography}

\vskip 0pt plus -1fil
\begin{IEEEbiography}
[{\includegraphics[width=1in,height=1.25in,clip,keepaspectratio]{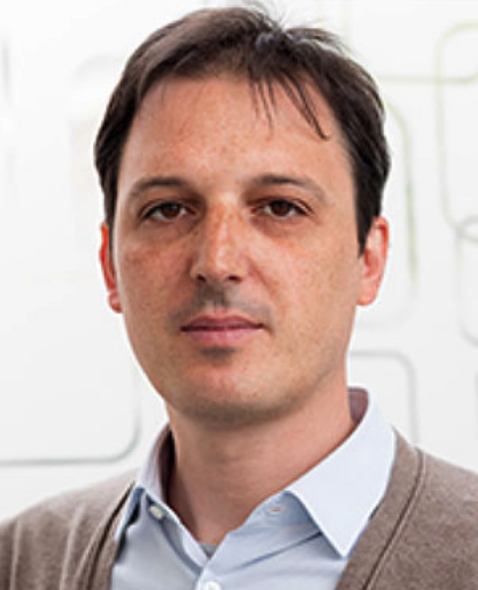}}]{Lorenzo Natale}Senior Member, IEEE) received the electronic engineering degree from University of Genova, Genoa, Italy, in 2000, and the Ph.D. degree in robotics from University of Genova, Genoa, Italy, in 2004. He was a Postdoctoral Researcher with the MIT Computer Science and Artificial Intelligence Laboratory and visiting Professor at the University of Manchester. He is currently a Senior Researcher with the Istituto Italiano di Tecnologia, Genoa, Italy, where he leads the Humanoid Sensing and Perception group. Lorenzo Natale was one of the main contributors to the design and development of the iCub platform, for which he led the development of the software architecture and middleware. His research interests range from vision and tactile sensing to software architectures for robotics. He served as General Chair of IEEE ARSO 2018 and Program Chair of ICDL-Epirob 2014 and HAI 2017. He is currently specialty Chief Editor for the Humanoid Robotics Section of Frontiers in Robotics and AI, Ellis Fellow and Core Faculty of the Ellis Genoa Unit.
\end{IEEEbiography}

\vskip 0pt plus -1fil
\begin{IEEEbiography}
[{\includegraphics[width=1in,height=1.25in,clip,keepaspectratio]{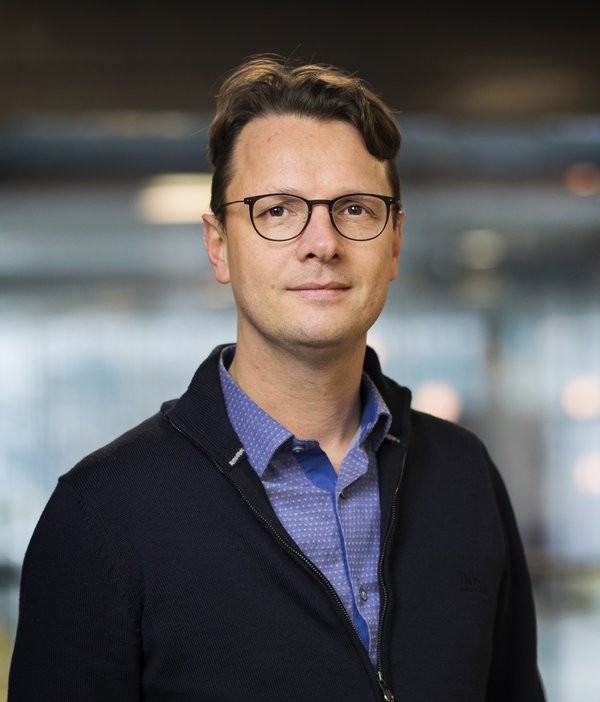}}]
{Raymond H. Cuijpers} graduated in Applied Physics at the TU/e in 1996. He received his PhD in Physics of Man from Utrecht University in 2000. He did a postdoc on the role of shape perception on human visuo-motor control at Erasmus MC Rotterdam. In 2004, he did a second postdoc at Radboud University Nijmegen. In 2008 he was appointed Assistant Professor at TU/e at the Human-Technology Interaction group. In 2014 he became associate professor in Cognitive Robotics and Human-Robot Interaction. His research interests include visual perception and visuo-motor control.
\end{IEEEbiography}

\end{document}